\pdfoutput=1
\documentclass[11pt]{article}

 \usepackage{acl}

\usepackage{times}
\usepackage{latexsym}
\usepackage{float}

\usepackage[T1]{fontenc}
\usepackage[utf8]{inputenc}

\usepackage{microtype}

\usepackage{inconsolata}

\usepackage{graphicx}

\usepackage{multirow}
\usepackage{microtype}
\usepackage{graphicx}
\usepackage{subfigure}
\usepackage{booktabs} % for professional tables
\usepackage{enumitem}
\usepackage{caption}
\usepackage{tcolorbox}

\usepackage{amsmath}
\usepackage{amsthm}
\newtheorem{definition}{Definition}

\title{Toward Human-AI Alignment in Large-Scale Multi-Player Games}
\author{
 \textbf{Sugandha Sharma\textsuperscript{1,2}},
\\
  \textbf{Guy Davidson\textsuperscript{1,3}},
 \textbf{Khimya Khetarpal\textsuperscript{1,4}},
 \textbf{Anssi Kanervisto\textsuperscript{1}},
 \textbf{Udit Arora \textsuperscript{1}},
\\
 \textbf{Katja Hofmann\textsuperscript{1}},
 \textbf{Ida Momennejad\textsuperscript{1}},
\\
\\
 \textsuperscript{1}Microsoft Research,
 \textsuperscript{2}Massachusetts Institute of Technology,\\
 \textsuperscript{3}New York University,
 \textsuperscript{4}Mila, University of Montreal 
\\
 \small{
   \textbf{Correspondence:} \href{mailto:susharma@mit.edu}{susharma@mit.edu}
 } 
 \\
  \small{
   \textit{Published in Wordplay @ ACL 2024, Association for Computational Linguistics}
 }
}
\begin{document}
\maketitle
\begin{abstract}
Achieving human-AI alignment in complex multi-agent games is crucial for creating trustworthy AI agents that enhance gameplay. We propose a method to evaluate this alignment using an interpretable task-sets framework, focusing on high-level behavioral tasks instead of low-level policies.
Our approach has three components. First, we analyze extensive human gameplay data from Xbox's Bleeding Edge (100K+ games), uncovering behavioral patterns in a complex task space. This task space serves as a basis set for a behavior manifold capturing interpretable axes: fight-flight, explore-exploit, and solo-multi-agent. Second, we train an AI agent to play Bleeding Edge using a Generative Pretrained Causal Transformer and measure its behavior. Third, we project human and AI gameplay to the proposed behavior manifold to compare and contrast. This allows us to interpret differences in policy as higher-level behavioral concepts,  e.g., we find that while human players exhibit variability in fight-flight and explore-exploit behavior, AI players tend towards uniformity. Furthermore, AI agents predominantly engage in solo play, while humans often engage in cooperative and competitive multi-agent patterns. These stark differences underscore the need for interpretable evaluation, design, and integration of AI in human-aligned applications. Our study advances the alignment discussion in AI and especially generative AI research, offering a measurable framework for interpretable human-agent alignment in multiplayer gaming.
\end{abstract}

\section{Introduction}
\label{intro}

\begin{figure*}[ht]
\begin{center}
\includegraphics[width=\textwidth]{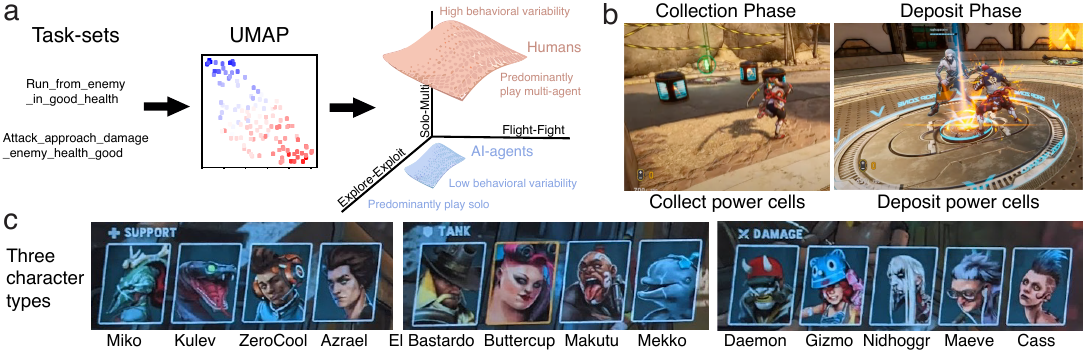}
%\vskip -0.1in
\caption{\textbf{Bleeding edge \textit{Power Collection} game mode.} (a) Analysis pipeline begins with task-sets used to extract the UMAP manifold embedding, interpreted to derive 3D human and AI behavioral manifold schematic. Humans highly vary in how they express fight-flight and explore-exploit behavior; they predominantly play in a multi-agent settings. AI agents exhibit low variability in fight-flight and explore-exploit behavior tending towards uniformity; they predominantly play solo. (b) Collection phase (\textit{left}) and Deposit phase (\textit{right}) in the Power Collection game mode. (c) Three character types (Support, Tank and Damage) with 13 possible characters in the game.}
\label{fig:bleeding_edge}
\end{center}
%\vskip -0.2in
\end{figure*}

Human-AI alignment is pivotal in generative AI research for several compelling reasons. First, as generative AI is increasingly integrated into various applications \cite{park2023generative, brynjolfsson2023generative}, ensuring alignment with human values and intentions becomes crucial to mitigate risks and enhance user trust \cite{sucholutsky2023getting, gabriel2020artificial}. Second, aligning AI systems with human behavior fosters more effective collaboration between humans and machines~\cite{chakraborti2018algorithms}, unlocking the potential for synergistic outcomes \cite{wynn2023learning, bobu2023aligning}. Third, in ethical considerations surrounding the deployment of generative models, alignment serves as a safeguard against unintended consequences and biases, promoting responsible AI development and deployment \cite{kenthapadi2023generative, weidinger2021ethical}. Thus, human-AI alignment in generative AI is essential for creating trustworthy, beneficial, and ethically sound AI applications that align with societal values and expectations.

The evaluation of human-agent alignment in rich observation and multi-agent environments poses a significant challenge \cite{leike2018scalable, ouyang2022training, wang2022self, wang2023aligning, burns2023weak}. In complex multi-agent video games, where each player faces a multitude of actions, this challenge involves finding the appropriate level of abstraction for a meaningful interpretation of human actions to evaluate artificial agents' alignment. 

In this work, we propose an interpretable approach towards human-AI alignment by introducing the ``Task-sets" framework, offering a means to abstract task sets from the environment. Task-sets \cite{sakai2008task} offer a higher level of abstraction compared to policies \cite{sutton1999policy, silver2014deterministic, lillicrap2015continuous, schulman2017proximal} and options \cite{sutton1999between, precup2000temporal, stolle2002learning, bacon2017option, khetarpal2020options} in reinforcement learning.

%For instance, a policy is a sequence of actions where an action can be \textit{turn left}, and a policy can be \textit{go left left left and then right}. An option is an abstraction of a policy, or a number of policies that take you from the same starting location to the same goal. However it's still at a lower level of abstraction than the concept of a task, which can have many different policies and options that lead to it. For instance an example of a task set is to decide whether a number is bigger or smaller than five. In this context there is a task domain, namely a number, and a task operation, which is to decide whether it is bigger or smaller than five, and an action by which the participant indicates their response. The number could be an actual number, it could be the number of doors on a street, or another entity. Likewise the response could be given by pressing a button for bigger than five and another button for smaller than five, or it could be by verbalizing it, or another modality of response. Regardless, a task set can capture the main point of what is being done, mainly to decide whether a magnitude is bigger or smaller than five and indicate the response. 

% Trying a different example
An agent's policy provides a low-level mapping from states to actions.
Suppose a person wants to get a snack from their kitchen. 
Their policy might indicate the precise sequence of steps (actions) to take to get from their living room to their kitchen. %which cupboard to open, and which snack to grab.
%Options offer one degree of abstraction over policies. 
The behavior described above might be split into two options: ``go-from-livingroom-to-kitchen'', and ``acquire-snack-in-kitchen''. 
The value of options lies in the fact that if a person is in their bedroom, they can execute a different first option (to arrive in the kitchen), and the same second option (acquiring a snack). 
The next day, however, at the office, their option's policy that facilitates acquiring a snack at home cannot help, as they need to navigate a different path, to a differently laid-out kitchen.
Task-sets describe a general higher degree of abstraction, ``walk to kitchen when hungry to find a snack,'' that is independent of the precise layout of the building (state space) and the steps needed to obtain the snacks (action space). 

\begin{tcolorbox}[colback=gray!10,colframe=black,boxrule=2pt,top=0.1in,bottom=0.1in,left=0.1in,right=0.1in]
\textbf{Key Definitions:}
\begin{itemize}[left=0pt,itemsep=0pt]
  \item \textbf{Task-set:} given a specific perceptual criteria (task domain) respond according to certain rules (task rules). 
  \item \textbf{Affordance:} when the criteria for performing a task set are met. Multiple task sets can be simultaneously afforded. 
  \item \textbf{Behavioral manifold:} dimensionality reduced space to which task set behavior and its spread  are projected.
  \item \textbf{Alignment on the manifold:} comparing human and agent spread along the dimensions of the behavioral manifold. 
\end{itemize}
%\vskip -0.3in
\end{tcolorbox}

We use the task-set abstraction to interpret differences between agents as higher-level behavioral concepts that transcend comparing changes in policies and options alone. This abstraction also allows comparing behavior across  %^This is the level of abstraction for tasks in this paper, that allows us to analyze far beyond simple notions of policies and options, and allows us to explore what agents do at various time scales from a few steps to many steps ahead i.e., different 
temporal scales \cite{monsell2003task, sakai2008task, collins2013cognitive, momennejad2013encoding, vaidya2022abstract}.
Task sets also facilitate compositionality.
In the example above, the ``walk to kitchen when hungry to find a snack'' task-set could also be used to grab a snack for a visitor (taking their preferences into account) or to modify one's own snack choice to account for healthiness. %or degree of hunger. % for instance you could still take the same domain and response approach above and merely change the task operation to be judging whether number is odd or even.
In Bleeding Edge, we use task-sets to abstract from actions and policies of players %in a game 
to higher-level notions, such as the dimensions of the behavior manifold (Fig.\ref{fig:bleeding_edge}a). This abstraction not only enables understanding of human cognitive processes, but also, fosters strong notions of transferability, both between agents and across environments, thus making it suitable for effective evaluation of alignment between AI agents and humans. 

Our \textbf{key contributions} are as follows: We propose an interpretable analysis of multi-scale behavior on different tasks by projecting them to behavior manifolds (Fig.\ref{fig:bleeding_edge}a) and evaluating human-AI alignment in this latent space. First, we analyze human gameplay data from the Xbox game Bleeding Edge ($\approx$ 100K games).
Our analysis uncovers human behavioral patterns in a complex task-set space.
We then interpret the agent's choices over which tasks to pursue at different moments in time as a behavior manifold capturing three interpretable axes: fight-flight, explore-exploit, and solo-multi-agent. Second, we train a proof of concept AI agent for gameplay using a Generative Pretrained Causal Transformer and measure its behavior using the same methods applied to the human data. Third, we project human and AI behavior to the same behavioral manifold and use the axes we defined to compare human-AI alignment. This three-fold analytical framework allows us to discern the extent of alignment between human and AI agent behaviors in a subspace defined by high-level and interpretable tasks, rather than policies. 

Our research, driven by these investigative avenues, pursues two-fold primary objectives. 1) We seek a nuanced understanding of human cognition and behavior in the realm of large-scale multiplayer video games. 2) We aspire to harness this understanding to advance AI for gameplay, by constructing, evaluating and training artifical agents for targeted behavior replication through player style identification. In this work we mainly focus on the evaluation framework for measuring alignment (illustrated through a proof of concept AI agent). This framework can be used for evaluating alignment of any autonomous decision making AI agent \cite{yao2022react, sharma2022map, shinn2023reflexion, Significant_Gravitas_AutoGPT, du2023guiding, wang2023voyager}. In summary, we provide a framework to evaluate human-AI alignment that could potentially be applied towards developing AI agents with superior alignment with humans.

\section{Bleeding Edge}
Bleeding Edge is a dynamic and engaging large-scale multiplayer online video game developed by Ninja Theory, blending fast-paced combat mechanics with team-based strategy. %Released in March 2020 for Microsoft Windows and Xbox One,
%The game introduces players to a captivating universe characterized by diverse characters.

\textbf{Gameplay:} 
Bleeding Edge is designed for 4v4 multiplayer battles. This means that two teams, each consisting of four players, participate in each game. Players engage in team-based battles featuring dynamic combat dynamics, including both melee (close-quarters physical engagements) and ranged (attacks from a distance) elements. %The multiplayer dynamics underscore the importance of teamwork, coordination, and communication in navigating the diverse and dynamic arenas in the game.

%Bleeding Edge is designed for 4v4 multiplayer battles. This means that two teams, each consisting of four players, participate in each game. Players engage in team-based battles featuring dynamic combat dynamics, including both melee and ranged elements. The primary objective varies across game modes, with teams strategically vying for control over points, completing tasks, or securing objectives to claim victory. The incorporation of both melee combat, involving close-quarters physical engagements, and ranged combat, which emphasizes attacks from a distance, ensures a diverse and challenging gameplay experience. The multiplayer dynamics underscore the importance of teamwork, coordination, and communication in navigating the diverse and dynamic arenas.

%The Power Collection game mode is a strategic multiplayer scenario designed for 4v4 engagements. 
\textbf{Power Collection game mode:} 
In this work, we restrict our analysis to the Power Collection game mode in Bleeding Edge. Central to this mode are two distinct phases: the Power Collection Phase and the Deposit Phase (Fig.\ref{fig:bleeding_edge}b). The objective revolves around the strategic acquisition of power cells (seeds) scattered across the game map (top-down view of a mini-map is visible to the players on the top-right corner of their screen). During the Power Collection Phase, teams are tasked with securing power cells positioned at specific locations, requiring meticulous planning and coordination. This phase introduces a dynamic interplay of risk and reward, as teams decide whether to focus on collecting cells nearer their base or venture farther into the map. The subsequent Deposit Phase involves transporting collected power cells to designated locations for scoring, further emphasizing the need for strategic decision-making and teamwork. Teams must defend their collected cells while attempting to disrupt opponents' efforts, contributing to the overall intensity and complexity of the gameplay. %This mode not only challenges players to adapt their tactics dynamically but also underscores the importance of effective communication and collaboration for success in this large-scale multiplayer setting. 

\textbf{Character selection and abilities:}
Players choose from a diverse roster of 13 characters, each with unique abilities and playstyles. Characters are classified into three categories (Fig.\ref{fig:bleeding_edge}c): Support, Tank, and Damage (see appendix \ref{app:character_types} for details). %Support characters focus on healing, buffs, and crowd control. Tanks absorb damage and protect teammates. Damage characters excel at eliminating opponents quickly. This allows for strategic team composition, promoting collaborative and strategic thinking (see appendix \ref{app:character_types} for details).

\section{AI Agent}

\begin{figure}[ht]
\begin{center}
% \centerline{\includegraphics[width=0.85\columnwidth]{icml2024/figures/
\includegraphics[width=\columnwidth]{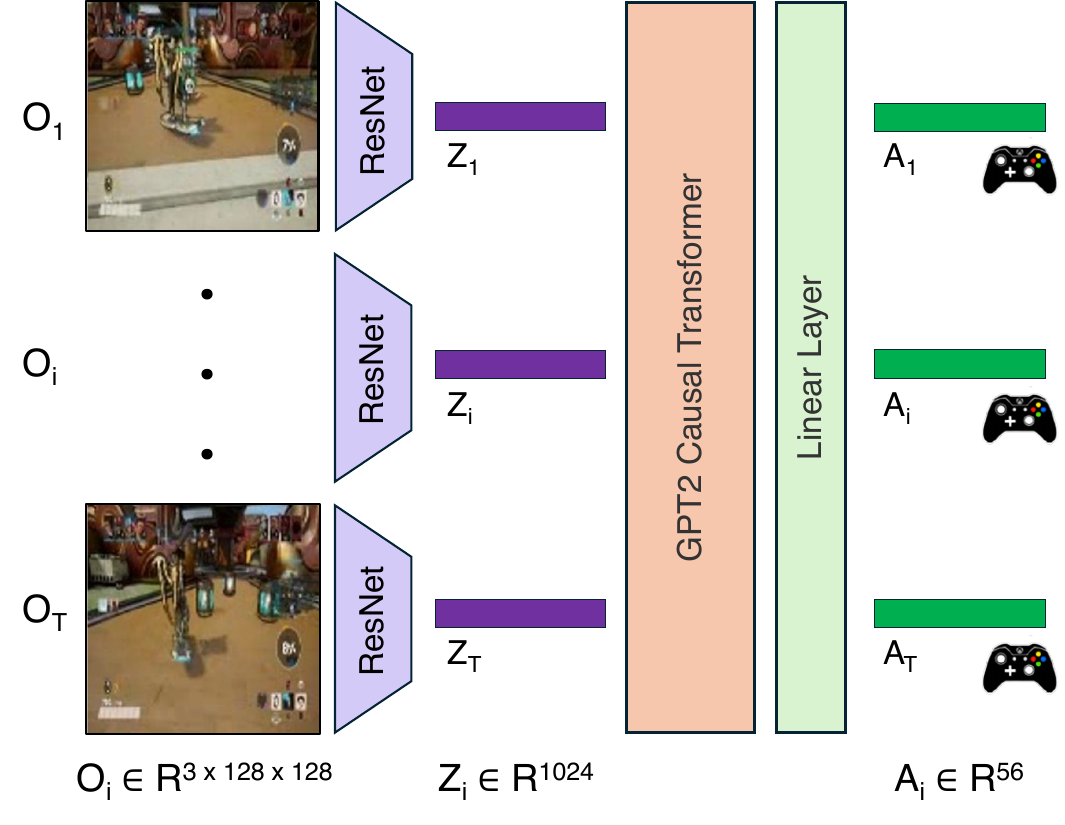}
%\vskip -0.1in
\caption{\textbf{AI agent architecture}. The architecture consists of a ResNet style encoder followed by a Causal Transformer. Model input is sequence of image observations ($O_i$), with sequence length $T$ and the model is trained to predict actions  ($A_i$).}
\label{fig:AI_agent}
\end{center}
%\vskip -0.15in
\end{figure}
%TODO: from Ida
% add a visualization of the three-D behavior manifold
% complex behavior can be turned into interpretable 

\begin{figure*}[ht]
\begin{center}
\includegraphics[width=\textwidth]{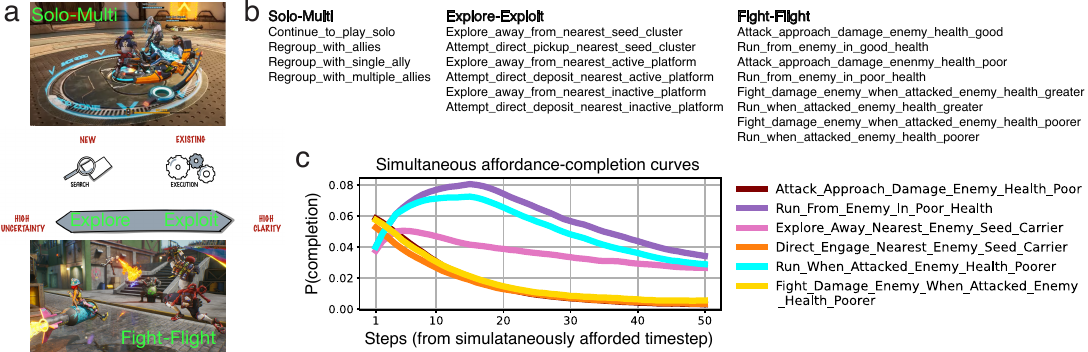}
%\vskip -0.1in
\caption{\textbf{Cognitive themes used for behavioral analysis. } (a) Ubiquitous cognitive themes in behaviors of various biological species used for analyzing gameplay dynamics of both human and AI agents. (b) Example task-sets defined for each of the cognitive themes. (c) Simultaneous affordance-completion curves for a subset of task-sets defined under the three cognitive themes.}
\label{fig-affordcomplcurves}
\end{center}
%\vskip -0.2in
\end{figure*}

We trained an %proof of concept 
AI agent for playing Bleeding Edge and measured its alignment with humans. The model, with $\sim$222M parameters, is a transformer based architecture. We frame human gameplay trajectories as sequences of image-action pairs, and optimize the transformer to predict the next action in the sequence given the previous images (Fig.\ref{fig:AI_agent}). %We use Behavior Cloning to train the model over a large human dataset playing Bleeding Edge. 

\textbf{Observation Encoder: } The model is trained on sequences of $T=128$ images where each image is reshaped to $128 \times 128 \times 3$ and then divided by 255 to ensure its value lies in range [0, 1]. A custom ResNet \cite{he2016deep} with 18.6M parameters is used to embed each image observation ($O \in R^{3\times128\times128}$) independently into a vector (details in \ref{app:observation_encoder}).  For each input image, the output of the encoder is a 1024D embedding.

%The first layer is a 2D convolutional network with kernels of shape $5 \times 5$, a stride of 3, and a padding of 1 and maps to 64 channel dimension. We then apply GELU \cite{hendrycks2016gaussian} activation. This is followed by 4 ConvNext \cite{liu2022convnet} and downsampling blocks. Each downsampling block applies group normalization and a convolution layer with kernel of shape $3 \times 3$, stride of 2, and padding of 1, doubling the number of channels. We again apply GELU activation followed by another 2D convolutional network with a kernel of shape $3 \times 3$, stride of $1 \times 3$ and padding of 0.

% UUU: is there a normalization layer?? that is in the convnext blocks
% UUU: how often are the observations sampled? e.g., sampled at 10Hz, 20Hz??

% The ResNet architecture consists of a 2D convoluational layer with kernels of size 5 x 5 followed by four blocks, each containing a convolutional, group normalization, convolutional, RELU activation and convolutional layers followed by a downsampling convolutional layer. 

\textbf{Transformer:}
The causal transformer (with $\sim$203M parameters) is applied on the image embeddings ($Z \in R^{1024}$). Specifically, a GPT2-like architecture \cite{radford2019language} from NanoGPT \cite{nanoGPT} was used containing 16 layers/blocks. Each attention layer has 8 heads with the action embeddings output of size 1024.
%UUU: Which hyperparameters does the transformer have? The hyperparams are 1) number of transf layers 2) embedding dim 3) bias when applying layernorm and feed forward MLP 4) Dropout value in MLP and Causal self attention 

\textbf{Action Decoder:} The final layer of the model consists of a linear layer that converts the output from the transformer (1024D) to match the dimensions of the action embedding (56D). The action space is an Xbox controller with two joysticks and 12 binary buttons. Each joystick is decomposed into $x$ and $y$ components leading to $4$ continuous values. Each of these continuous values are discretized by binning them to 11 bins, such that the model predicts the logits over which bin is the most likely. This leads to $12 + 4 \times 11 = 56$ dimensional action output ($A \in R^{56}$).

%The output of the transformer is fed into the action decoder consisting of a GELU activation followed by a linear layer to the embedding from the transformer's output dimension to action dimension i.e. 56. The first 12 dimensions correspond to the button actions. The next four sets of 11 dimensions correspond to $x$ component of left joystick, $y$ component of left joystick, $x$ component of right joystick and $y$ components of right joystick. 

% The leads to $12+4=16$ total action dimensions. Vocabulary is assigned such that each action dimension has its own unique tokens, 11 for each joystick dimension, and two for buttons, leading to a total action vocab size of, $Va := 68 = (4 \times 11) + (12 \times 2)$.

%UUU: should we lable the action decoder in the agent diagram in Figure 3? 

\textbf{Data and Training:}
We use Behavior cloning \cite{pomerleau1991efficient} for training. For buttons we use the binary cross entropy loss with logits, and for the joysticks we use the cross entropy loss for each component. The total loss is computed as the sum of the losses for buttons and each of the joystick's components. The training data was sampled from a dataset consisting of 57,661 full human gameplay videos where each video corresponds to continuous gameplay by one player, resulting in 1,707,997,180 video frames ($\sim 1.8$ billion time steps) and 7907.4 hours of human gameplay. The training took 6 days on 16 GPUs (V100s) and the model was trained for $\sim 72000$ steps (details in \ref{app:training}).

\textbf{AI Rollouts:}
We generated $600$ rollouts of $1$ min each with the above model by first picking a number of random game situations from the dataset. These are then filtered down to the desired characters e.g., Daemon, ZeroCool, and Makutu. (see appendix \ref{ai_rollouts} for details).

\begin{figure*}[ht]
\begin{center}
\includegraphics[width=\textwidth]{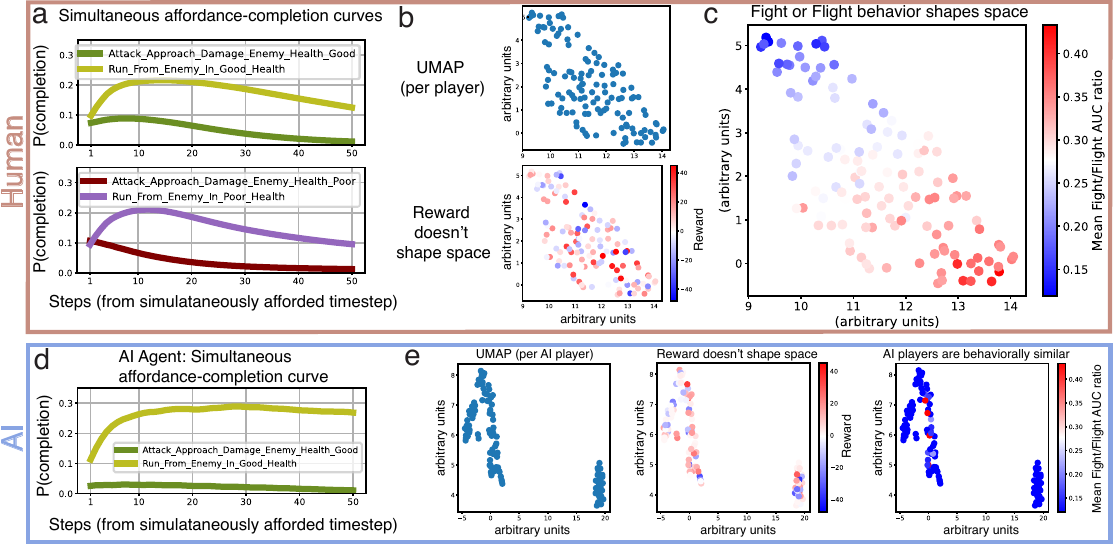}
%\vskip -0.15in
\caption{\textbf{Fight-Flight analysis results. }(a) Simultaneous affordance-completion curves for two representative pairs of fight-flight task-sets from human gameplay data. (b) \textit{Top:} An unsupervised 2D UMAP (Uniform Manifold Approximation and Projection) embedding of 123 human players averaged across 637 games. Each point represents one human player. \textit{Bottom:} Human UMAP colored by reward. (c) Human UMAP colored by fight-flight behavior. (d) Simultaneous affordance-completion curves for a representative pair of fight-flight task-set from AI gameplay data. (e) \textit{Left:} UMAP embeding of 116 AI players averaged across 116 games. Each point represents one AI player. \textit{Middle:} AI UMAP colored by reward. \textit{Right:} AI UMAP colored by fight-flight behavior.}
\label{fig-fight_flight}
\end{center}
%\vskip -0.2in
\end{figure*}
%playing the Daemon character

\section{Behavioral analysis of gameplay data}
\label{sec:task-sets}

\subsection{Task-sets}

We introduce the Task-sets framework for an interpretable analysis of human behavior from 100,000 games of Bleeding Edge (Power Collection game mode).  %Task-sets, as defined in cognitive sciences \cite{sakai2008task}, are configurations of cognitive processes actively maintained for subsequent task performance. A task-set comprises features, attention, and actions, forming a generalized notion of tasks that encompasses environmental features such as color of an object or number of objects in agent's field of view. Agents select specific features, assess them by attending to them (e.g., identifying if color is red or if the number is odd), and produce a sequence of actions to complete the task-set. A task-set is afforded, within this framework, when the conditions for performing a task-set are met.
We here formalize the definition of task-sets in conjunction with affordance and completion conditions over features of the (latent) state. %This involves extracting a set of features from the game state at each time step. \textbf{Affordance conditions} identify when a task-set can be performed or executed, giving the agent the choice of whether or not to engage in the task-set. \textbf{Completion conditions} determine if an agent successfully performed a task-set by choosing to engage. 

\begin{definition}[Task-Set]
A task-set comprises of extracting a set of features from the game state at each time step on which affordance and completion conditions are determined. \textit{Affordance conditions} identify when a task-set can be performed or executed, giving the agent the choice of whether or not to engage in the task-set. \textit{Completion conditions} determine if an agent successfully performed a task-set by choosing to engage. A task-set is said to be afforded when its affordance condition is met, and said to be completed when it's completion condition is met.
\end{definition}
%\vskip -0.1in

This compositional approach to agent behavior, expressed as the composition of various task sets, underscores the flexibility and adaptability of the task-set framework in capturing agent behavior. To illustrate, consider the following task-sets % with corresponding affordance and completion conditions are given below 
(see the full list of all task-sets in Fig.\ref{fig-task-sets-list}): 

\textit{Run\_From\_Enemy\_In\_Good\_Health} \newline
\textbf{Affordance condition:} the nearest enemy is (a) within 2100 distance units of our character, and (b) has above 50\% of their health remaining. \newline
\textbf{Completion condition:} our character is (a) moving away from the nearest enemy, and (b) that nearest enemy is within 3500 distance units.

\textit{Attack\_Approach\_Damage\_Enemy\_Health\_Good} \newline
\textbf{Affordance condition:} the closest enemy to our character is (a) within 2100 distance units from the ego character, (b) they have over 50\% of their health remaining, and (c) the ego character is moving toward them. \newline
\textbf{Completion condition:} the ego character either dealt damage or was credited with a kill on this timestep.

%As shown above, each task-set is defined in conjunction with an affordance and completion condition. 
We note that future work could consider using automatically learned task-sets similar to skill learning \cite{wang2023voyager, khetarpal2021temporally}. However, in the scope of this work, we adhere to the programmer-specified definition of task sets based on our understanding of the game and the analysis of game play data, a deliberate choice aimed at illustrating the advantages inherent in this framework without being limited by the quality of learned task sets.

%It is noteworthy that task sets can be defined in two ways: they can either be hand-defined or learned automatically. In the context of this work, we adhere to the manual definition of task sets, a deliberate choice aimed at illustrating the advantages inherent in this framework. %This deliberate decision to hand-define task sets underscores the potential insights gained into human behavior and highlights the potential associated advantage of building agents that exhibit robust generalization across diverse environments.

\textbf{Cognitive themes:}
In our analytical framework, we employ three distinct cognitive themes that are ubiquitously observed in the behaviors of various biological species (Fig.\ref{fig-affordcomplcurves}a). They are: 1)``Fight-Flight" for shedding light on the decision-making processes associated with confrontation and evasion, 2) ``Explore-Exploit" is employed to discern the strategic balance between exploration and exploitation strategies within the game environment, and 3)``Solo-Multi-agent play" is used for understanding the interplay between individual and collaborative player behaviors. The task-sets defined for each of these cognitive themes are shown in Fig.\ref{fig-affordcomplcurves}b (complete definitions in \ref{app:task-set_definitions}). 

%Thus, our analysis incorporates ubiquitous themes observed in the behaviors of various biological species. This holistic approach, rooted in cognitive science, ensures a nuanced and comprehensive examination of player actions and interactions, offering valuable insights into the cognitive processes underpinning the gameplay dynamics. 

\begin{figure*}[ht]
\begin{center}
\includegraphics[width=\textwidth]{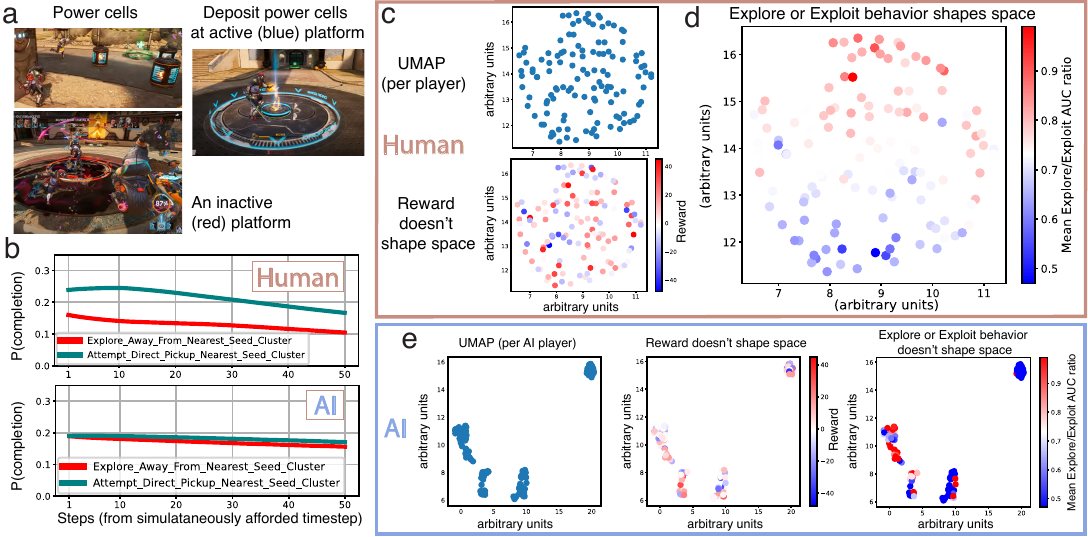}
%\vskip -0.15in
\caption{\textbf{Explore-Exploit analysis results.} (a) Goal-directed navigation based task-set illustrations. (b) Simultaneous affordance-completion curves for a representative pair of explore-exploit task-set from human \textit{top} and AI (\textit{bottom}) gameplay data. (c) \textit{Top:} An unsupervised 2D UMAP embedding of 123 human players averaged across 637 games. Each point represents one human player. \textit{Bottom:} Human UMAP colored by reward. (d) Human UMAP colored by explore-exploit behavior. (e) \textit{Left:} UMAP embeding of 116 AI players averaged across 116 games. Each point represents one AI player. \textit{Middle:} AI UMAP colored by reward. \textit{Right:} AI UMAP colored by explore-exploit behavior.}
\label{fig-explore_exploit}
\end{center}
%\vskip -0.2in
\end{figure*}

\subsection{Simultaneous affordance-completion analysis}

% TODO: from Ida
% Panel (c): edit caption to make the take-away clear
% how can these planes be used to infer alignment
\begin{figure*}[ht]
\begin{center}
%\centerline{\includegraphics[trim=0cm 0.04cm 0cm 0cm,clip, width=0.95\textwidth]{icml2024/figures/figure5.pdf}}
\includegraphics[width=\textwidth]{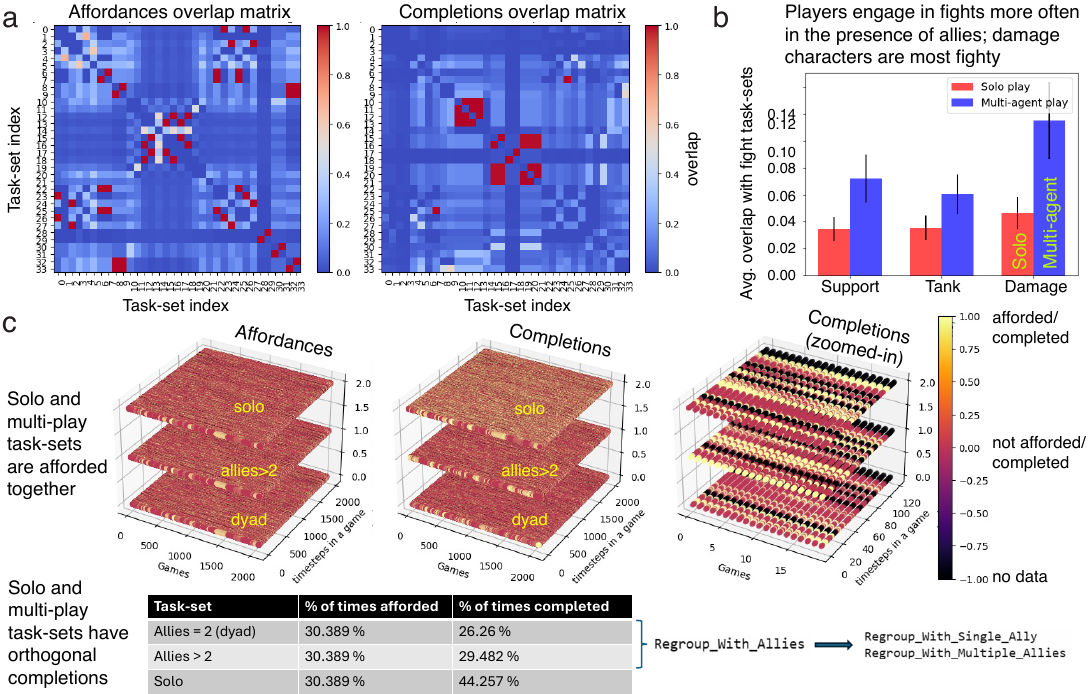}
%\vskip -0.15in
\caption{\textbf{Differences in human play-style across character types.} (a) \textit{Left:} Average task-set affordances (\textit{left}) and completions (\textit{right}) overlap matrix for Damage characters. (b) Overlap of solo-multi-agent task-sets with fight task-sets averaged across all characters within each character type (Fig.\ref{fig:bleeding_edge}c). (c) \textit{Left, Mid: } Affordances and completions respectively of solo-multi-agent task-sets (Fig.\ref{fig-affordcomplcurves}b, Left) plotted for a subset of timesteps: $\sim 2000$ time steps per game for $2059$ games played by the Daemon character. \textit{Right: } Zoomed in version of the completions plot showing orthogonal completions across the three task sets. \textit{Bottom:} Table listing the average $\%$ of times each of the solo-multi-agent task sets were afforded/completed across all the characters and games.}
\label{fig-diff-playstyle}
\end{center}
%\vskip -0.2in
\end{figure*}

We systematically analyze agent behavior across three aforementioned different axes that highlight meaningful variation: fight-flight, explore-exploit, and solo- vs. multi-agent play. For each axis, we identified a collection of task-sets that capture behavior along this axis, and conduct the following analysis:
% \newline

\textbf{Identification of Afforded Timesteps:} For each timestep of every game in our dataset, we identify states at which all relevant task-sets were afforded to the agent.

\textbf{Evaluation of Task-Set Completion:} For each of the simultaneously afforded task-sets, we examine whether it was completed before it was afforded again to the agent. If completed, we record all future timesteps of potential completions before the next affordance.

\textbf{Probability of Completion Computation:} We aggregate this completion data across all simultaneous completion timesteps, and use it to compute the probability of completion of each of the afforded task-sets. Specifically, for each simultaneously afforded task-set, we compute the probability of completion at time $t+x$, given a simultaneous affordance at time $t$: \\
$P($completion at $t + x |$ simul. afford. at $t) $ 
\\
$ = \frac{\text{\# completions of current task-set } x \text{ steps after affordance} }{ \text{total \# of timesteps the task-sets were simultaneously afforded }} $
\\
% (i) \textbf{Identification of Afforded Timesteps:} For each timestep of every game in our dataset, we identify states at which all relevant task-sets were afforded to the agent. 
% \newline
% (ii) \textbf{Evaluation of Task-Set Completion:} For each of the simultaneously afforded task-sets, we examine whether it was completed before it was afforded again to the agent. If completed, we record all future timesteps of potential completions before the next affordance.
% \newline
% (iii) \textbf{Probability of Completion Computation:} We aggregate this completion data across all simultaneous completion timesteps, and use it to compute the probability of completion of each of the afforded task-sets. Specifically, for each simultaneously afforded task-set, we compute the probability of completion at time $t+x$, given a simultaneous affordance at time $t$:

% $P($completion at $t + x |$ simultaneous affordance at $t) $ 

% $ = \frac{\text{\# completions of current task-set } x \text{ steps after affordance} }{ \text{total \# of timesteps the task-sets were simultaneously afforded }} $

That is, for each future step $x$, we add up the completion counts from different time-steps in which the given task-set combination was afforded, and divide by the total number of observations of the combination.
Plotting this data produces the simultaneous affordance-completion curves for the task-sets (e.g., Fig.\ref{fig-affordcomplcurves}c).

\section{Results}
\subsection{Fight-Flight}
\label{sec:fight-flight}

\textbf{Hypothesis: } Players vary in how they express the fight or flight behavior while playing the same character.

To test this hypothesis, four pairs of task-sets (Fig.\ref{fig-affordcomplcurves}b, right) that examine fight (attacking) or flight (running away) behavior were used for the analysis. Each pair has a different affordance criteria (appendix \ref{app:fight-flight}). For each set of simultaneous affordances, we compute completion probabilities for both task-sets, and compare their completion probability curves as shown in Fig.\ref{fig-fight_flight}a. Note that the flight task-sets have a higher completion probability than the fight task-sets. For human gameplay data, we limited the analysis to the players who had played $3+$ games controlling the Daemon character. Hence, the analysis was conducted for 123 players who played a total of 637 games with Daemon. 

For each pair of task-sets, we generate $9$ features: Area under the curve (AUC), Max, and Argmax of ‘fight’ task-set curve; AUC, Max, and Argmax of ‘flight’ task-set curve; and the ratio between each of the features, dividing fight / flight, resulting in $4\times9 = 36$ interpretable features per player. We use these features to produce an unsupervised 2D embedding of each human player through UMAP (Uniform Manifold Approximation and Projection) \cite{mcinnes2018umap} as shown in Fig.\ref{fig-fight_flight}b, top. Here each point corresponds to one human player leading to a total of 123 points on the manifold. The data for each player is averaged across all the games (3 or more) played by this player. When the manifold is colored by the reward (score) received by the players, we find that the reward doesn't shape the embedding space as shown in Fig.\ref{fig-fight_flight}b, bottom. However, when the manifold is colored by fight/flight AUC ratio, we find that the human behavioral manifold embedding space is shaped by fight or flight behavior as shown in Fig.\ref{fig-fight_flight}c. Although on an absolute scale, all players are flighty, relatively some players are more fighty (red points) than others. 

With the same analysis on the AI agent gameplay data: Fig. \ref{fig-fight_flight}d shows a representative simultaneous affordance-completion curve showing that the AI has a very low completion probability for fight task-sets. Fig.\ref{fig-fight_flight}e, left shows the UMAP embedding of 116 AI players averaged across 116 games, while playing Daemon. Here each point corresponds to one AI player leading to a total of 116 points on the manifold. Fig.\ref{fig-fight_flight}e, middle shows the manifold colored by reward received, indicating that the reward doesn't shape the AI embedding space. Fig.\ref{fig-fight_flight}e, right shows that when colored by fight/flight AUC ratio, almost all AI players are behaviorally similar, i.e., almost all of them are flighty. This is in contrast to the human players who are behaviorally different on a relative scale with some of them being more fighty and some being more flighty (Fig.\ref{fig-fight_flight}c). 

\textbf{Conclusion: } We find that while human players vary in how they express the fight or flight behavior while playing the same character, AI players do not. Apart from Daemon, we also ran the above analysis for other characters in Fig.\ref{fig:bleeding_edge}c and drew the same conclusion irrespective of the character. %We also investigated the change in human player strategies when the same player plays a different character (Fig.\ref{fig-character_switch}). %From the limited data we had, 
%We found that on switching characters, a greater $\%$ of players $\approx57\%$ switched to flight behavior and only $\approx44\%$ switched to fight behavior reinforcing the tendency of human players towards flight behavior on an absolute scale. 

\subsection{Explore-Exploit}

\textbf{Hypothesis: } Players vary in how they express the explore or exploit behavior while playing the same character.

To test this hypothesis we defined three pairs of goal-directed navigation based task-sets listed in Fig.\ref{fig-affordcomplcurves}b, middle and shown in Fig.\ref{fig-explore_exploit}a. Exploitation prioritizes immediate rewards by efficiently achieving the objectives in a direct and goal-oriented manner. %For instance, moving directly towards the nearest cluster to collect power cells, going directly to the nearest active platform for depositing them as quickly as possible, or going to the nearest inactive platform and waiting for it to become active. 
Similar to the fight-flight analysis above (section \ref{sec:fight-flight}), we compute the simultaneous affordance-completion curves for each pair of explore-exploit task-sets for human players as well as AI agents. Fig.\ref{fig-explore_exploit}b,top shows that humans are exploiters - the exploit task-set has a higher completion probability than the explore task-set. Next, we compute the UMAP embedding as shown in Fig.\ref{fig-explore_exploit}c,top. Again, we find that the reward doesn't shape human behavioral manifold embedding space (Fig.\ref{fig-explore_exploit}c, bottom), however, when colored by explore/exploit AUC ratio, the embedding space is shaped by explore or exploit behavior (Fig.\ref{fig-explore_exploit}d). Although on an absolute scale, all players are exploiters, relatively some players explore more (red points) than others.

%For human gameplay data, we limited the analysis to the players who had played $3+$ games controlling the Daemon character. Hence, the analysis was conducted for 123 players who played a total of 637 games with Daemon. 

%Here each point corresponds to one human player leading to a total of 123 points on the manifold. The data for each player is averaged across all the games (3 or more) played by this player.

%Next, we use the features from the simultaneous-completion curves to produce an unsupervised 2D embedding of each human player through 

%Here each point corresponds to one AI player leading to a total of 116 points on the manifold.

We ran the same analysis on the AI agent gameplay data. Fig. \ref{fig-explore_exploit}b,bottom shows that the AI has almost equal completion probability for the explore and exploit task-set. Fig. \ref{fig-fight_flight}e,left shows the UMAP embedding of AI players, while playing Daemon. Fig.\ref{fig-fight_flight}e,middle shows the manifold colored by reward received, indicating that the reward doesn't shape the AI embedding space. Fig.\ref{fig-fight_flight}e,right shows that when colored by explore/exploit AUC ratio, the AI embedding space is not shaped by explore-exploit behavior either. This indicates that although AI players show some diversity in exploring vs exploiting, both of these groups share similar feature representations, making it difficult to differentiate between them based on explore-exploit behavior. In other words, AI agents exhibit overlapping characteristics between exploration and exploitation behavior and can't be grouped based on it.  This is in contrast to the human players who are behaviorally different on a relative scale with some of them being more of explorers and some being more of exploiters (Fig.\ref{fig-explore_exploit}d). 

\textbf{Conclusion: } We find that while human players clearly vary in how they express explore or exploit behavior when playing the same character, we can't differentiate between AI players based on this behavior. This conclusion stands across characters.\footnote{We find that the same analysis for different characters in Fig.\ref{fig:bleeding_edge}c results in the same findings.}

%Apart from Daemon, 
%We also ran the above analysis for various other characters in Fig.\ref{fig:bleeding_edge}c and reached the same conclusion irrespective of the character played. %SS:TODOapp 

\subsection{Solo-Multi-agent play}
To identify differences in play-style across character types, we compute the overlap in task-set affordances and completions (Fig.\ref{fig-diff-playstyle}a) for each character type (Fig.\ref{fig-overlap_affordance}a, \ref{fig-overlap_completions}a). On analysing the difference between these overlap matrices (Fig.\ref{fig-overlap_affordance}b, \ref{fig-overlap_completions}b), we find that the Tank characters are the most suited to carrying power cells (Fig.\ref{fig-overlap_tanks_powercells}), Support characters to healing (Fig.\ref{fig-support_healing}), and Damage characters to dealing damage (Fig.\ref{fig-damage_dealingdamage}). From the completions overlap matrices, we extract the overlap of completions of solo-multi-agent game play task sets (Fig.\ref{fig-diff-playstyle}b, left) with the fight task-sets in the human gameplay data. We find that irrespective of the character type, all human players engage in fights more often in the presence of allies relative to when they are playing solo. Additionally, we find that the Damage characters are the most fighty. 

Besides, we find that all the solo-mutli-agent task-sets are afforded simultaneously as is evident by the three identical affordance planes shown in Fig.\ref{fig-diff-playstyle}c,left. However, all of them have orthogonal completions illustrated by the three distinct completion planes (see Fig.\ref{fig-diff-playstyle}c, middle and right). The table in the figure summarizes the fraction of times the three task-sets are afforded and completed (relative to affordances) in the game. Although all of them are simultaneously afforded, their completions are orthogonal, leading to completion fractions that sum up to $100$ for the three task sets. When playing the Daemon character, humans play in cooperation with their allies $\approx 56\%$ of the time, and play solo only $\approx 44 \%$ of the time.  

Table \ref{table:human_solo_multi} (in appendix) lists the percentage ($\%$) of time spent in solo vs multi-agent game play by human players when playing different characters %from the three character types 
(Fig.\ref{fig:bleeding_edge}c). Irrespective of the character played, humans spend a significantly greater amount of time playing cooperatively with their allies in multi-agent settings relative to solo game play. We subject the AI game play data to the same analysis for a single randomly chosen character for each character type with results summarized in Table \ref{table:AI_human_solo_multi}. We find that in contrast to humans, AI agents spend majority of game time ($\approx 70 \%$) playing solo.

\section{Discussion} We propose a Task-sets-based framework for (i) understanding human behavior in large-scale multiplayer games, and (ii) assessing the alignment of AI agents with humans. We apply this framework to examine the alignment of a proof of concept GPT-based AI agent trained to play Bleeding Edge. Through simultaneous affordance-completion analysis of task-sets, we examine interpretable behavioral axes, allowing for richer comparisons than what policy differences alone allow for. %enable

Our analysis of the human data shows that the task-sets capture meaningful factors of variation in human behavior along the three behavioral axes. %fight-flight, explore-exploit, and solo-multi-agent tendencies. %suggesting they are a useful way to measure alignment along these axes. 
While humans show substantial differences along these dimensions, data from our AI agent does not mirror the variations observed in human behavior. 
We take this as evidence that these AI agents are not aligned with humans. % and hence is not aligned with humans. 
%which suggests that cooperation between humans and these AI agents might be challenging. 
%In particular, these agents may exhibit behaviors not easily anticipated by human collaborators, potentially leading to a less engaging and immersive gameplay experience for human players. 
See \ref{app:future_impact} for future work, broader impact and societal implications. 

% when playing with these AI agents.
% since the AI agents we evaluate may be incompatible% with human players, exhibiting behaviors not easily anticipated by humans. This could lead to a less engaging and immersive gameplay experience for human players. % when playing with these AI agents.

% Bibliography entries for the entire Anthology, followed by custom entries
%\bibliography{anthology,custom}
% Custom bibliography entries only
\bibliography{custom}

%%%%%%%%%%%%%%%%%%%%%%%%%%%%%%%%%%%%%%%%%%%%%%%%%%%%%%%%%%%%%%%%%%%%%%%%%%%%%%%
%%%%%%%%%%%%%%%%%%%%%%%%%%%%%%%%%%%%%%%%%%%%%%%%%%%%%%%%%%%%%%%%%%%%%%%%%%%%%%%
% APPENDIX
%%%%%%%%%%%%%%%%%%%%%%%%%%%%%%%%%%%%%%%%%%%%%%%%%%%%%%%%%%%%%%%%%%%%%%%%%%%%%%%
%%%%%%%%%%%%%%%%%%%%%%%%%%%%%%%%%%%%%%%%%%%%%%%%%%%%%%%%%%%%%%%%%%%%%%%%%%%%%%%
\newpage
\appendix
\onecolumn
\section{Appendix}
\renewcommand\thefigure{\thesection.\arabic{figure}} 

%You can have as much text here as you want. The main body must be at most $8$ pages long.
%For the final version, one more page can be added.

%The $\mathtt{\backslash onecolumn}$ command above can be kept in place if you prefer a one-column appendix, or can be removed if you prefer a two-column appendix.  Apart from this possible change, the style (font size, spacing, margins, page numbering, etc.) should be kept the same as the main body.

\begin{figure*}[ht]
\begin{center}
\includegraphics[width=\textwidth]{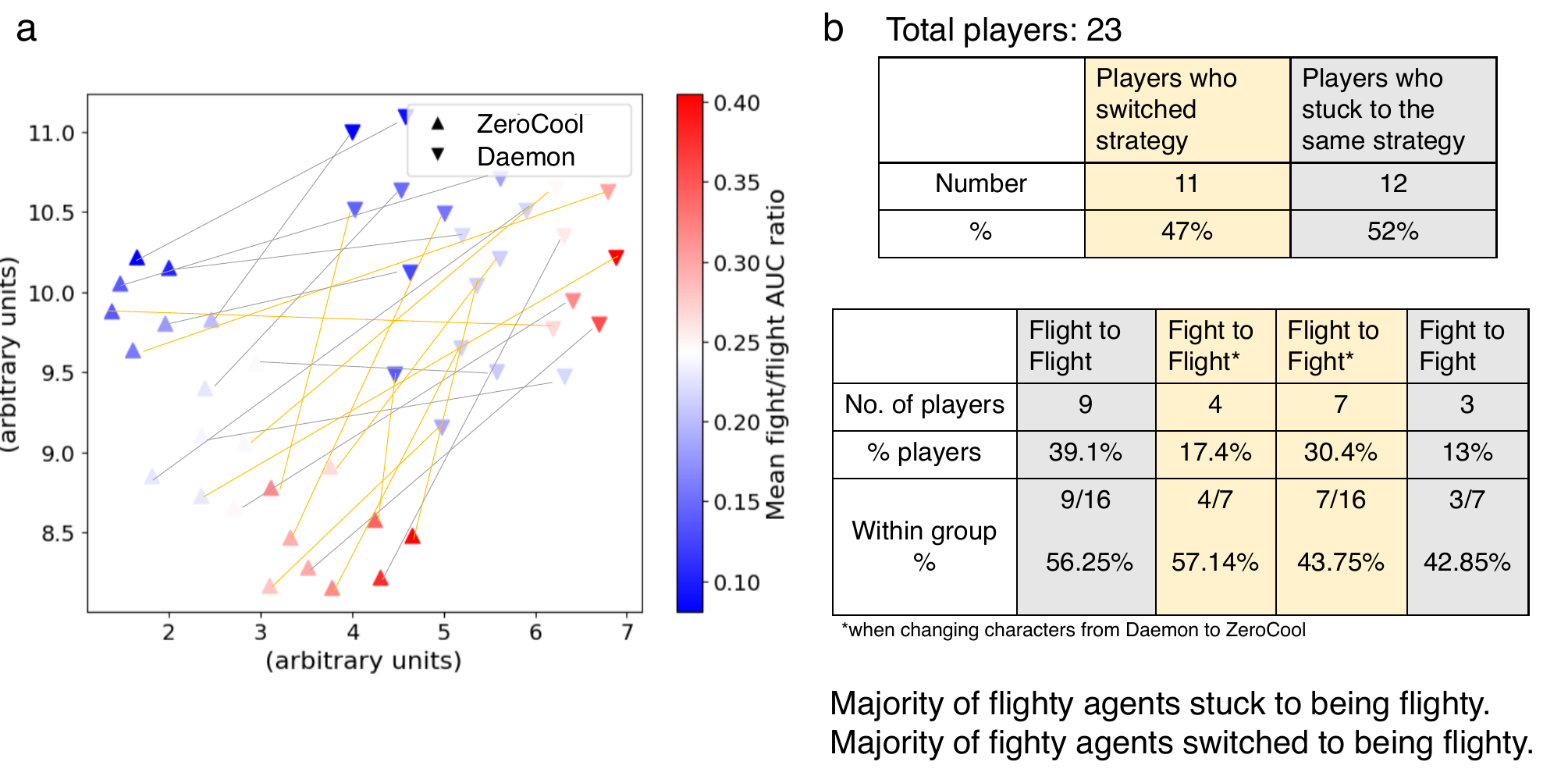}
\caption{\textbf{Change in player strategies when changing characters.} \textit{Left:} A manifold showing the change in strategies when switching between ZeroCool and Daemon characters. \textit{Right: } Summary of results from the switching dynamics analysis of all players who played $3+$ games with both ZeroCool and Daemon. Out of the players who switched strategies, a greater $\%$ of players switched to Flight relative to switching to Fight, and out of the players who stuck to the same strategy, a greater $\%$ were flighty rather than fighty.}
\label{fig-character_switch}
\end{center}
\end{figure*}

\begin{figure*}[ht]
\begin{center}
\includegraphics[width=\textwidth]{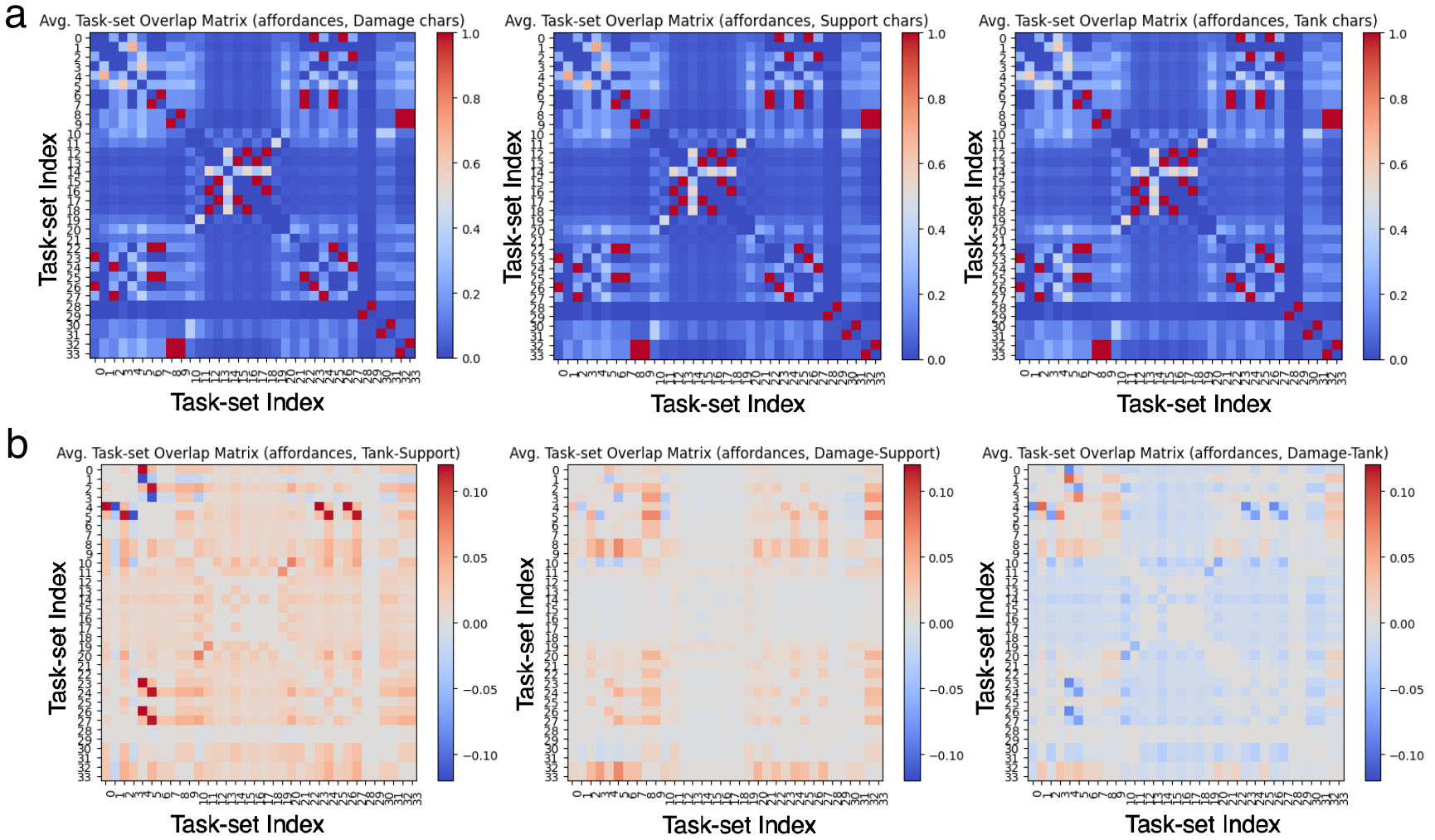}
\caption{\textbf{Overlap in task-set affordances.} (a) Avg. task-set affordance overlap matrix for \textit{Left:} Damage characters, \textit{Middle:} Support characters, \textit{Right:} Tank characters. (b) Difference in the matrices in (a). \textit{Left:} Tank-Support, \textit{Left:} Damage-Support, \textit{Left:} Damage-Tank.}
\label{fig-overlap_affordance}
\end{center}
\end{figure*}

\begin{figure*}[ht]
\begin{center}
\includegraphics[width=0.85\textwidth]{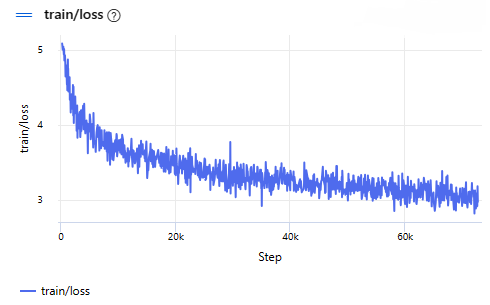}
\caption{\textbf{AI agent loss curve}. The decreasing loss curve indicates that over time, the AI agent improves its performance and makes action predictions that are increasingly closer to the ground truth. The smooth, steady decrease in the loss suggests that the AI agent is learning effectively and converging towards an optimal solution.}
\label{fig-task-sets-list}
\end{center}
\end{figure*}

\begin{figure*}[ht]
\begin{center}
\includegraphics[width=\textwidth]{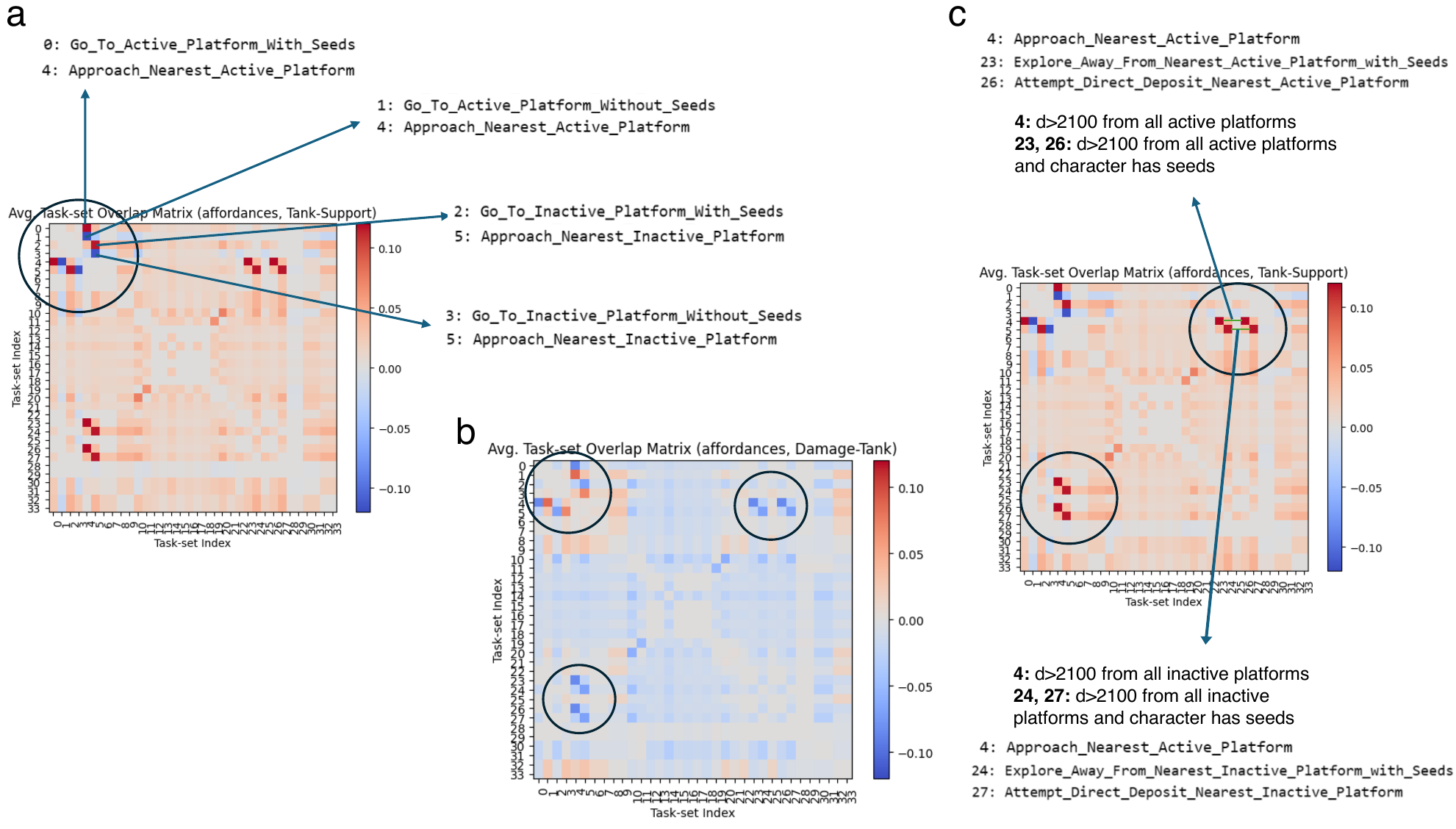}
\caption{\textbf{Analysis of overlap in task-set affordances shows that Tanks are the most suited to carrying power cells.} (a, c) Tanks have a higher overlap with task-sets that involve going to a platform with power cells showing that Tanks are more likely to go to platform with seeds than Supports. (b) Tanks are more likely to go to a platform with seeds than Damage characters. }
\label{fig-overlap_tanks_powercells}
\end{center}
\end{figure*}

\begin{figure*}[ht]
\begin{center}
\includegraphics[width=\textwidth]{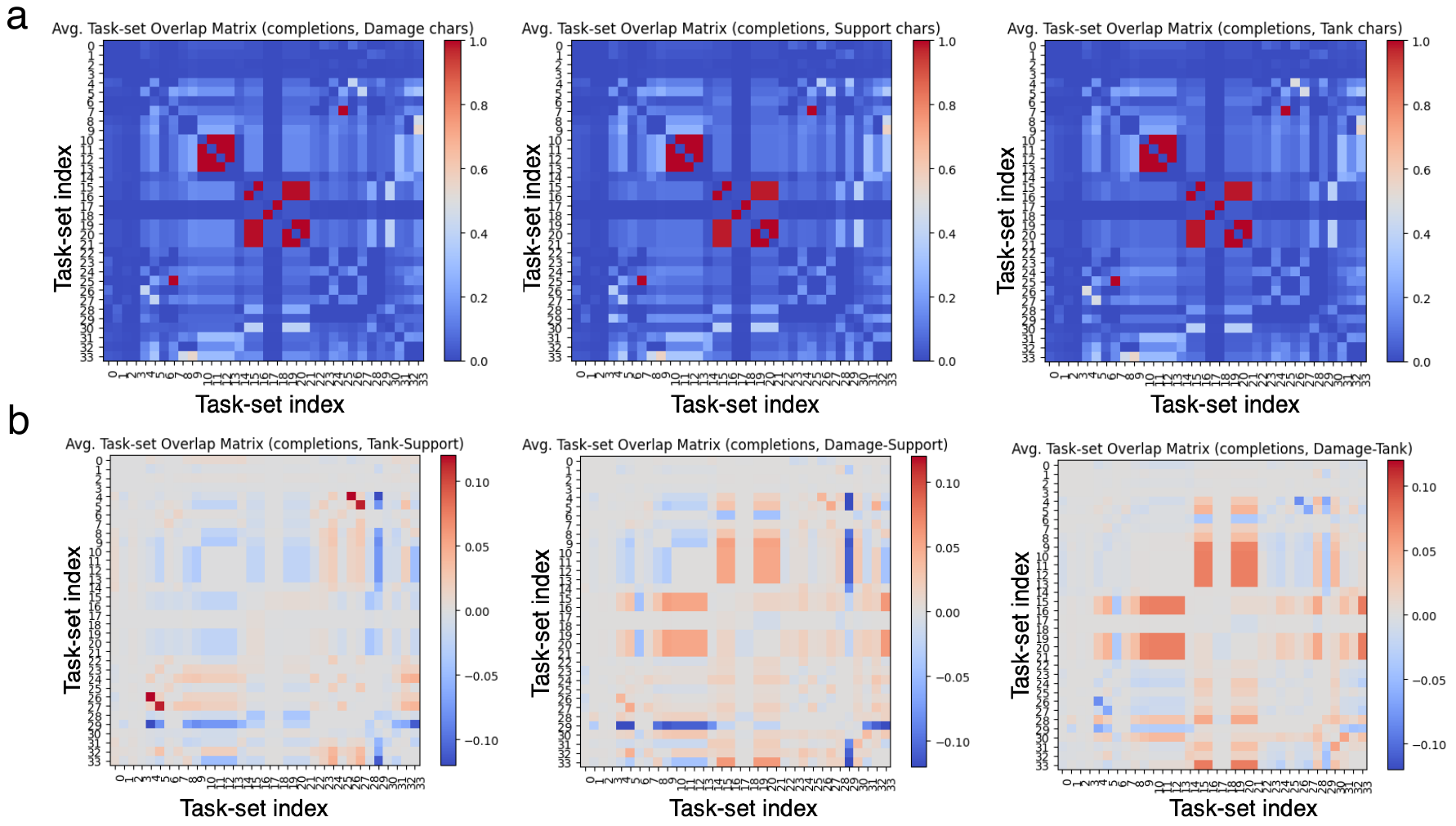}
\caption{\textbf{Overlap in task-set completions.} (a) Avg. task-set completions overlap matrix for \textit{Left:} Damage characters, \textit{Middle:} Support characters, \textit{Right:} Tank characters. (b) Difference in the matrices in (a). \textit{Left:} Tank-Support, \textit{Left:} Damage-Support, \textit{Left:} Damage-Tank.}
\label{fig-overlap_completions}
\end{center}
\end{figure*}

\begin{figure*}[ht]
\begin{center}
\includegraphics[width=\textwidth]{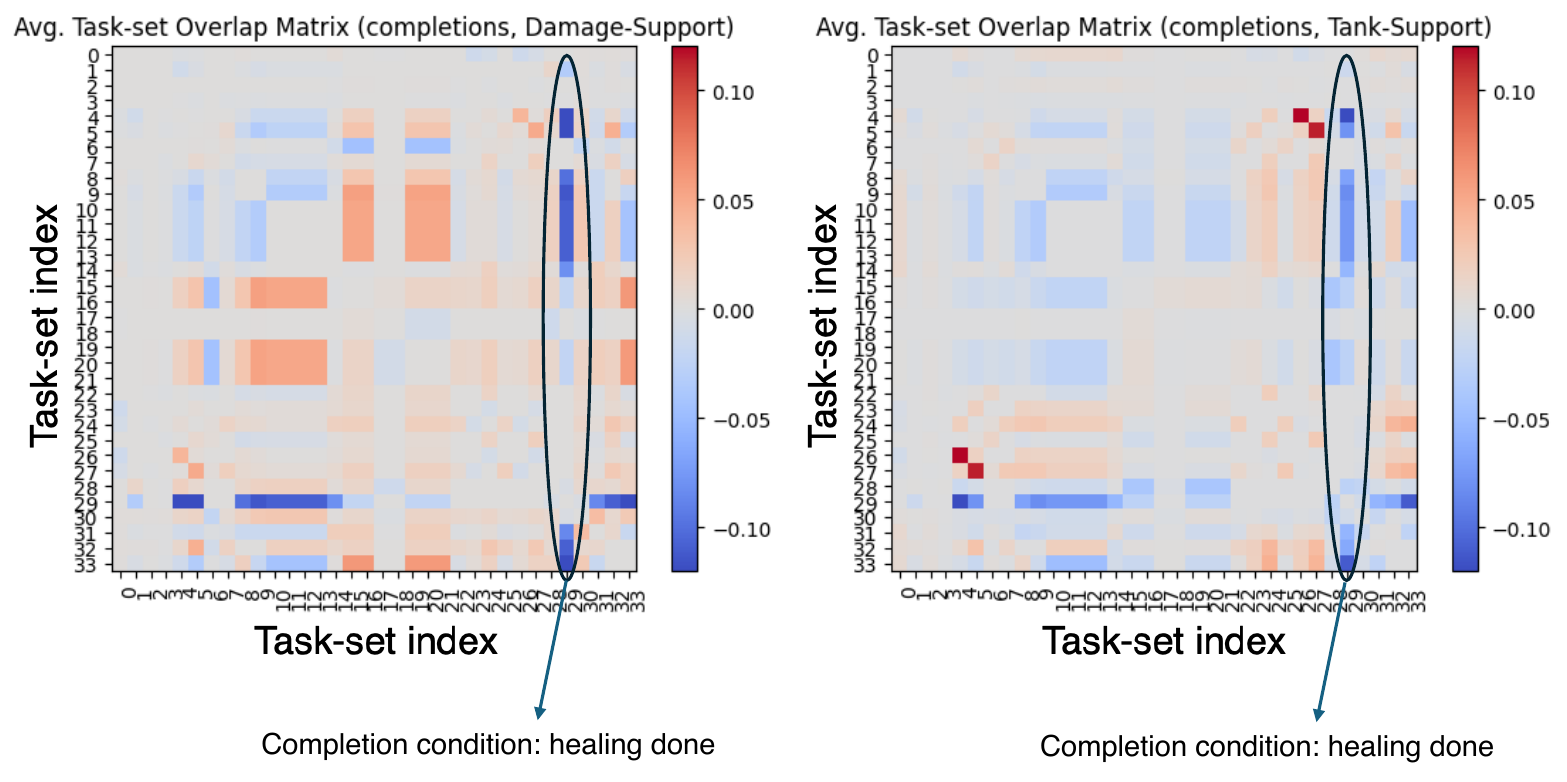}
\caption{\textbf{Analysis of overlap in task-set completions shows that Supports are the most suited to healing.} Difference in completions matrices of Damage and Support as well as Tank and Support characters shows that Supports have a higher overlap with all the task-sets that have a completion condition that includes healing.}
\label{fig-support_healing}
\end{center}
\end{figure*}

\begin{figure*}[ht]
\begin{center}
\includegraphics[width=\textwidth]{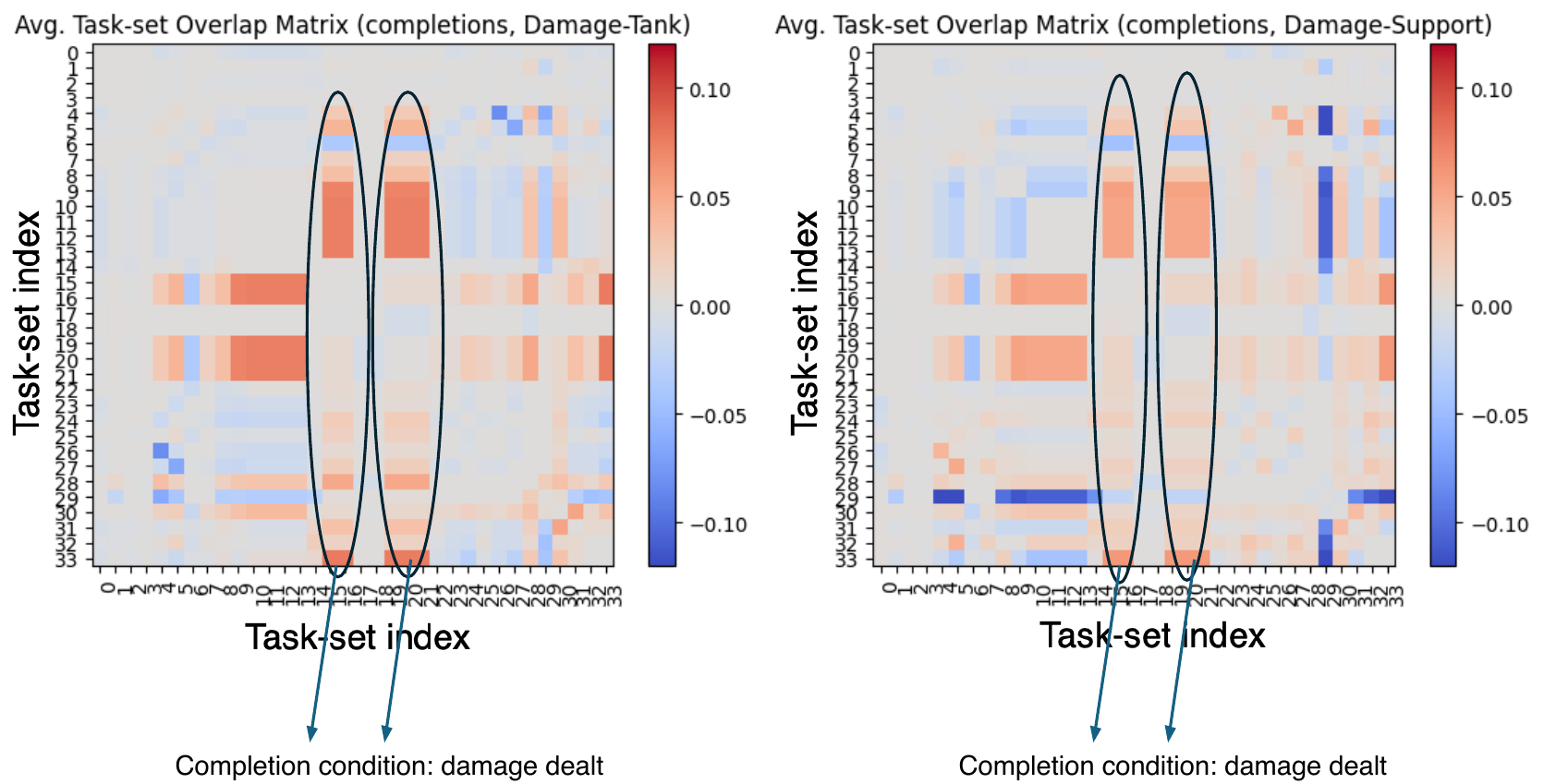}
\caption{\textbf{Analysis of overlap in task-set completions shows that Damage characters are the most suited to dealing damage.} Difference in completions matrices of Damage and Tank as well as Damage and Support characters shows that Damage characters have a higher overlap with all the task-sets that have a completion condition that includes dealing damage.}
\label{fig-damage_dealingdamage}
\end{center}
\end{figure*}

\begin{figure*}[ht]
\begin{center}
\includegraphics[width=\textwidth]{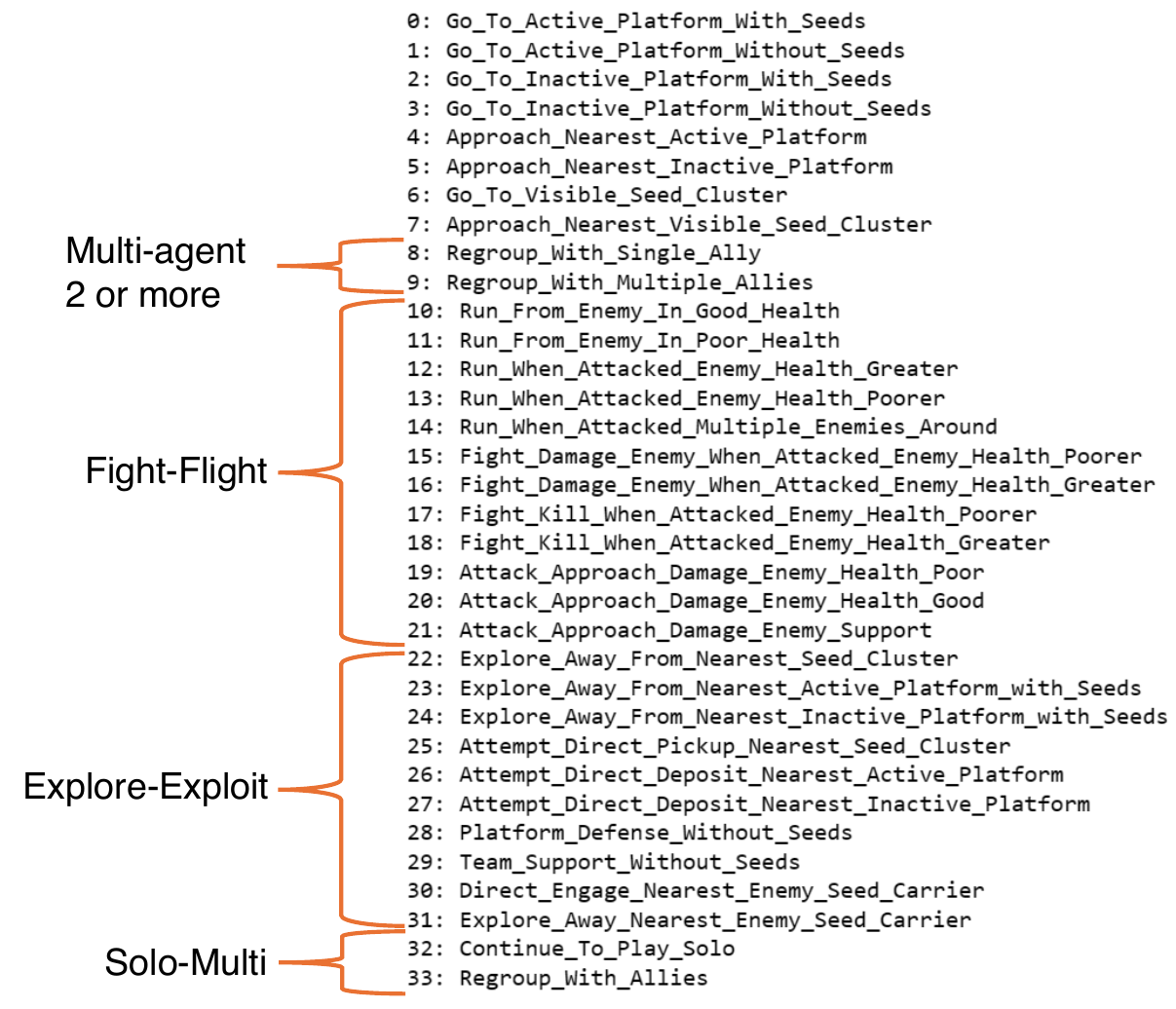}
\caption{\textbf{Task-sets}. List of all task-sets implemented for the analysis presented in this paper. Each of these was implemented as a routine in python}
\label{fig-task-sets-list}
\end{center}
\end{figure*}

\subsection{Acknowledgements}

We would like to acknowledge Tabish Rashid at Microsoft Research for support and guidance on the AI agent. We would also like to acknowledge the Ninja Theory team (that developed Bleeding Edge) for their feedback during the course of this project.  

% \subsection{Software and data}
% The source code for the model presented in this paper, a data sample, and the analysis code can be made available upon acceptance (for the camera-ready version).

\subsection{Why Bleeding Edge}

Although there are multiple games that we could have chosen for the work presented in this paper, we opted to work with Bleeding Edge since we had access to rich human behavioral data from human game play in this game that was easy to work with and process, made possible by our current affiliations. This data was used to train the AI agents presented in the paper as well as for human behavioral analysis through the task-sets framework in order to study human-AI alignment. 
%with Microsoft

\subsection{AI Agent details}

\subsubsection{Training}
\label{app:training}

For this work, we train the agent on less than one epoch for computational time and memory complexity reasons. While training we add data augmentation to the video frames following \citet{baker2022video}.
% Please add the following required packages to your document preamble:
% \usepackage{multirow}
\begin{table}[H]
\caption{Hyperparameters for AI agent training}
\centering
\begin{tabular}{|l|l|}
\hline
\textbf{Parameter}     & \textbf{Value}       \\ \hline
Steps                  & 72,000               \\ \hline
Learning Rate          & 0.0001               \\ \hline
Warmup Steps           & 1000                 \\ \hline
Optimizer              & AdamW                \\ \hline
Optimizer weight decay & 0.0001               \\ \hline
Batch Size             & 12                   \\ \hline
Sequence length        & 128                  \\ \hline
L2 Gradient Clipping   & 1.0                  \\ \hline
\end{tabular}
\end{table}
The learning rate is scheduled by the following function:
\begin{equation}
    lr = min((steps + 1)/warmupSteps, 1)
\end{equation}

\subsubsection{Observation Encoder}
\label{app:observation_encoder}
The model is trained on sequences of $T=128$ images where each image is reshaped to $128 \times 128 \times 3$ and then divided by 255 to ensure its value lies in range [0, 1]. A custom ResNet \cite{he2016deep} with 18.6M parameters is used to embed each image observation ($O \in R^{3\times128\times128}$) independently into a vector. The first layer is a 2D convolutional network with kernels of shape $5 \times 5$, a stride of 3, and a padding of 1 and maps to 64 channel dimension. We then apply GELU \cite{hendrycks2016gaussian} activation. This is followed by 4 ConvNext \cite{liu2022convnet} and downsampling blocks. Each downsampling block applies group normalization and a convolution layer with kernel of shape $3 \times 3$, stride of 2, and padding of 1, doubling the number of channels. We again apply GELU activation followed by another 2D convolutional network with a kernel of shape $3 \times 3$, stride of $1 \times 3$ and padding of 0. For each input image, the output of the encoder is a 1024D embedding.

% add table with hyperparameters
% add a few lines for anything else thats missing

\subsection{Task-set definitions}
\label{app:task-set_definitions}

\subsubsection{Fight-Flight}
\label{app:fight-flight}

We define four pairs of task-sets, one each for fight and flight:

\begin{enumerate}
    \item Absolute enemy health $> 50\%$: \\
    \textbf{Affordance condition:} the nearest enemy is within 2100 distance units from the ego character, has above $50\%$ (‘good’) of their health remaining and the ego character is moving toward them.\\
        \begin{enumerate}
            \item \textit{Fight: Attack\_Approach\_Damage\_Enemy\_Health\_Good} \\
            \textbf{Completion condition:} the ego character dealt damage on this timestep.\\
            
            \item \textit{Flight: Run\_From\_Enemy\_In\_Good\_Health} \\
            \textbf{Completion condition:} the ego character is moving away from the nearest enemy, and the nearest enemy is within 3500 distance units from the ego character.\\
        \end{enumerate}

    \item Absolute enemy health $< 50\%$: \\
    \textbf{Affordance condition:} the nearest enemy is within 2100 distance units from the ego character, and has below $50\%$ (‘poor’) of their health remaining.\\
        \begin{enumerate}
            \item \textit{Fight: Attack\_Approach\_Damage\_Enemy\_Health\_Poor} \\
            \textbf{Completion condition:} the ego character dealt damage on this timestep.\\
            
            \item \textit{Flight: Run\_From\_Enemy\_In\_Poor\_Health} \\
            \textbf{Completion condition:} the ego character is moving away from the nearest enemy, and the nearest enemy is within 3500 distance units from the ego character.\\
        \end{enumerate}

    \item Enemy health $>$ Player health: \\
    \textbf{Affordance condition:} the nearest enemy is within 2100 distance units, they have a larger (‘greater’) $\%$ of their health remaining than the ego character, and the ego character took damage on this timestep.\\
        \begin{enumerate}
            \item \textit{Fight: Fight\_Damage\_Enemy\_When\_Attacked\_Enemy\_Health\_Greater} \\
            \textbf{Completion condition:} the ego character dealt damage on this timestep.\\
            
            \item \textit{Flight: Run\_When\_Attacked\_Enemy\_Health\_Greater} \\
            \textbf{Completion condition:} the ego character is moving away from the nearest enemy, and the nearest enemy is within 3500 distance units from the ego character.\\
        \end{enumerate}

    \item Enemy health $<$ Player health: \\
    \textbf{Affordance condition:} the nearest enemy is within 2100 distance units, they have a lower (‘poorer’) $\%$ of their health remaining than the ego character, and the ego character took damage on this timestep.\\

        \begin{enumerate}
            \item \textit{Fight: Fight\_Damage\_Enemy\_When\_Attacked\_Enemy\_Health\_Poorer} \\
            \textbf{Completion condition:} the ego character dealt damage on this timestep.\\
            
            \item \textit{Flight: Run\_When\_Attacked\_Enemy\_Health\_Poorer} \\
            \textbf{Completion condition:} the ego character is moving away from the nearest enemy, and the nearest enemy is within 3500 distance units from the ego character.\\
        \end{enumerate}

\end{enumerate}

\subsubsection{Explore-Exploit}
\label{app:explore-exploit}

We define three pairs of task-sets, one each for explore and exploit:

\begin{enumerate}
    \item Seed collection strategy: \\
    \textbf{Affordance condition:} there exists at least one visible seed cluster, ego character distance $>$ 2100 from all visible seed clusters.\\
        \begin{enumerate}
            \item \textit{Exploit: Attempt\_Direct\_Pickup\_Nearest\_Seed\_Cluster    } \\
            \textbf{Completion condition:} there exists at least one visible seed cluster, the ego character is moving towards the nearest seed cluster.\\
            
            \item \textit{Explore: Explore\_Away\_From\_Nearest\_Seed\_Cluster } \\
            \textbf{Completion condition:} the ego character is moving away from the nearest seed cluster.\\
        \end{enumerate}

    \item Deposit strategy (relevant in deposit phase): \\
    \textbf{Affordance condition:} the ego character has seeds (number of seeds $> 0$) and distance of ego character $> 2100$ units from all active platforms .\\
        \begin{enumerate}
            \item \textit{Exploit: Attempt\_Direct\_Deposit\_Nearest\_Active\_Platform} \\
            \textbf{Completion condition:} the ego character has seeds (number of seeds $> 0$), and is moving towards the nearest active platform.\\
            
            \item \textit{Explore: Explore\_Away\_From\_Nearest\_Active\_Platform\_with\_Seeds} \\
            \textbf{Completion condition:} the ego character is moving away from the nearest active platform.\\
        \end{enumerate}

    \item Deposit strategy (relevant in collection phase): \\
    \textbf{Affordance condition:} the ego character has seeds (number of seeds $> 0$), ego character distance $> 2100$ units from all inactive platforms.\\
        \begin{enumerate}
            \item \textit{Exploit: Attempt\_Direct\_Deposit\_Nearest\_Inactive\_Platform} \\
            \textbf{Completion condition:} the ego character has seeds (number of seeds $> 0$), and is moving towards the nearest inactive platform.\\
            
            \item \textit{Explore: Explore\_Away\_From\_Nearest\_Inactive\_Platform\_with\_Seeds} \\
            \textbf{Completion condition:} the ego character is moving away from the nearest inactive platform.\\
        \end{enumerate}

\end{enumerate}

\subsubsection{Solo-Multi}

We define task-sets to study solo vs multi-agent game play dynamics.

\textbf{Affordance condition:} no single teammates within a distance of 3500

\begin{enumerate}

    \item \textit{Solo: Continue\_To\_Play\_Solo} \\
    \textbf{Completion condition:} distance from nearest teammate $>$ 2100.
    
    \item \textit{Regroup: Regroup\_With\_Allies} \\
    \textbf{Completion condition:} distance from nearest teammate $<$ 2100.
    
    \item \textit{Diad: Regroup\_With\_Single\_Ally} \\
    \textbf{Completion condition:} distance from nearest teammate $<$ 2100 and only one teammate is present within this distance range.

    \item \textit{Multi-agent: Regroup\_With\_Multiple\_Allies} \\
    \textbf{Completion condition:} distance from multiple (more than one) teammates $<$ 2100.

\end{enumerate}

\label{app:solo-multi}

\subsection{Character types}
\label{app:character_types}

When playing Bleeding Edge, players select their character, from a diverse roster of 13 characters, each with a unique set of abilities and playstyles. The characters are classified into three main categories (Fig.\ref{fig:bleeding_edge}c): 

\begin{enumerate}
    \item Support: Possess healing abilities, buffs, crowd control, or other utility tools. They excel at keeping their allies alive, providing additional damage or defense boosts, and disrupting the enemy team. 
    \item Tank: Durable and resilient, capable of soaking up large amounts of damage and protecting their teammates. They have high health pools and often possess abilities that allow them to mitigate or redirect damage away from their more fragile allies. 
    \item Damage: Excel at engaging in fights and eliminating opponents quickly. They tend to have lower health pools and may require support or protection from Tank characters to survive in prolonged engagements.
\end{enumerate}

The varied selection of characters allows for strategic team composition, encouraging players to tailor their choices to complement their team's overall strategy. This diversity promotes collaborative and strategic thinking as players work together to capitalize on each character's strengths. 

\subsection{AI Rollouts}
\label{ai_rollouts}

For each game, we initialize the game at that state and let the AI agent play from that state onwards for a duration of $1$ minute. Except for the AI agent, all other players in the game are non-players who aren't controlled and simply continue to repeat the latest action that the corresponding human player acted during the real human gameplay (e.g., if the player was moving forward when the game was originally recorded at that state, they will continue moving forward). They do not respond to the AI agent’s actions, however, their features get affected based on AI agent’s actions. For instance, their health decreases when the AI agent attacks them. We assume that each rollout is a different AI player given the stochasticity e.g., due to sampling the action from the output probability distribution.

\begin{table}[ht]
\begin{center}
\includegraphics[width=0.7\columnwidth]{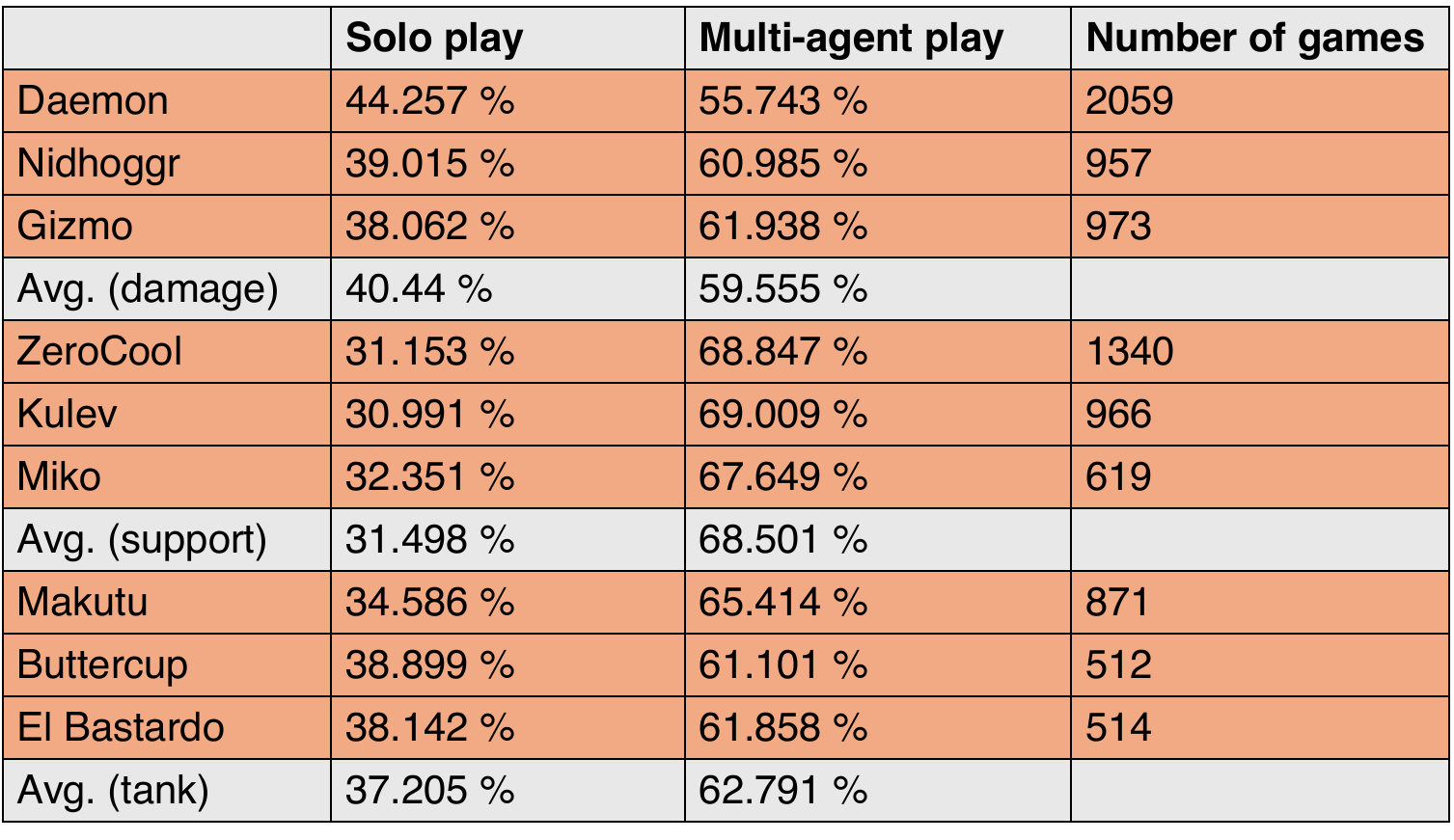}
%\vskip -0.15in
\caption{Percentage of time spent by human players playing solo (player playing alone) is substantially less than that spent playing with their allies (multi-agent: 2 allies or more playing together) for the characters in three character types (Fig.\ref{fig:bleeding_edge}c) averaged across all games played by the given character.}
\label{table:human_solo_multi}
\end{center}
%\vskip -0.3in
\end{table}
%($\%$)

\begin{table}[ht]
% \captionsetup{font=scriptsize} 
\begin{center}
% \centerline{\includegraphics[width=0.95\columnwidth]{icml2024/figures/solo_multi_human_agent.pdf}}
\includegraphics[width=0.7\columnwidth]{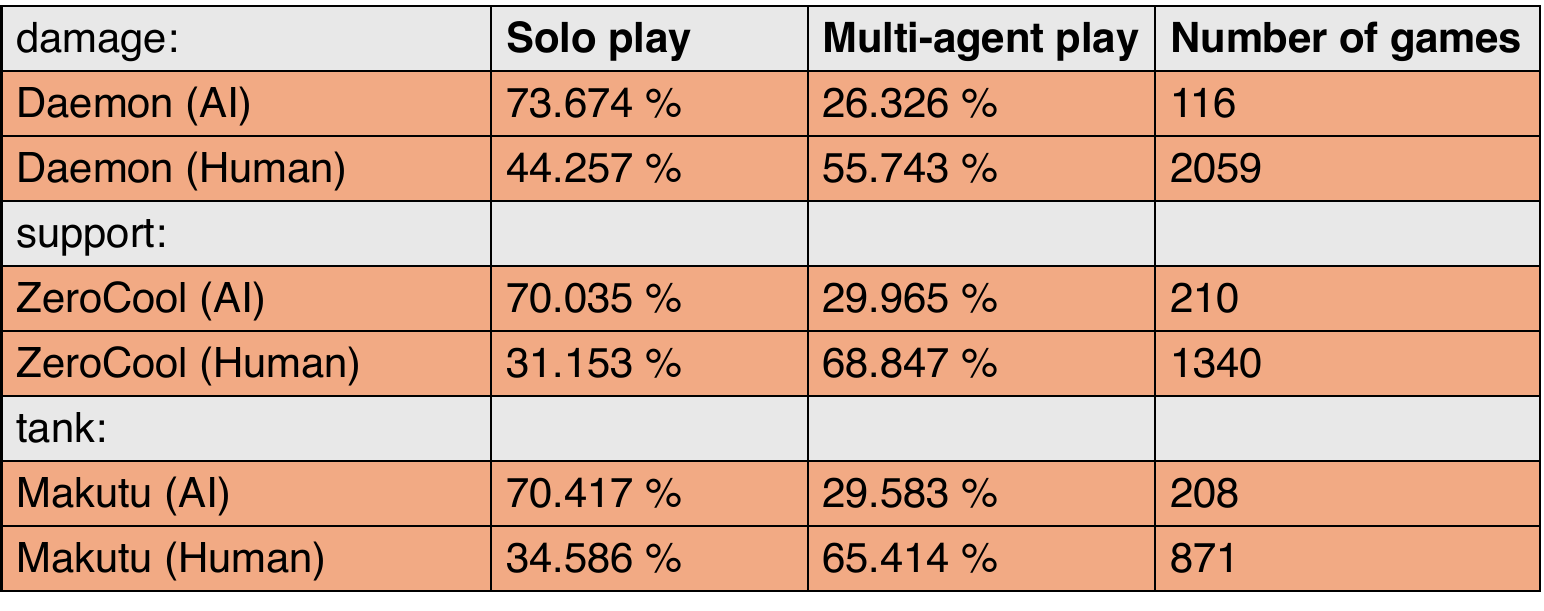}
%\vskip -0.15in
\caption{Percentage of time spent by AI vs Human players playing alone (solo) and with their allies (multi-agent) %: 2 allies or more playing together) 
for one character from each of the three character types (Fig. \ref{fig:bleeding_edge}c) averaged across all games played by a given character. AI predominantly plays solo.}
\label{table:AI_human_solo_multi}
\end{center}
%\vskip -0.3in
\end{table}

\begin{figure*}[ht]
\begin{center}
\includegraphics[width=\textwidth]{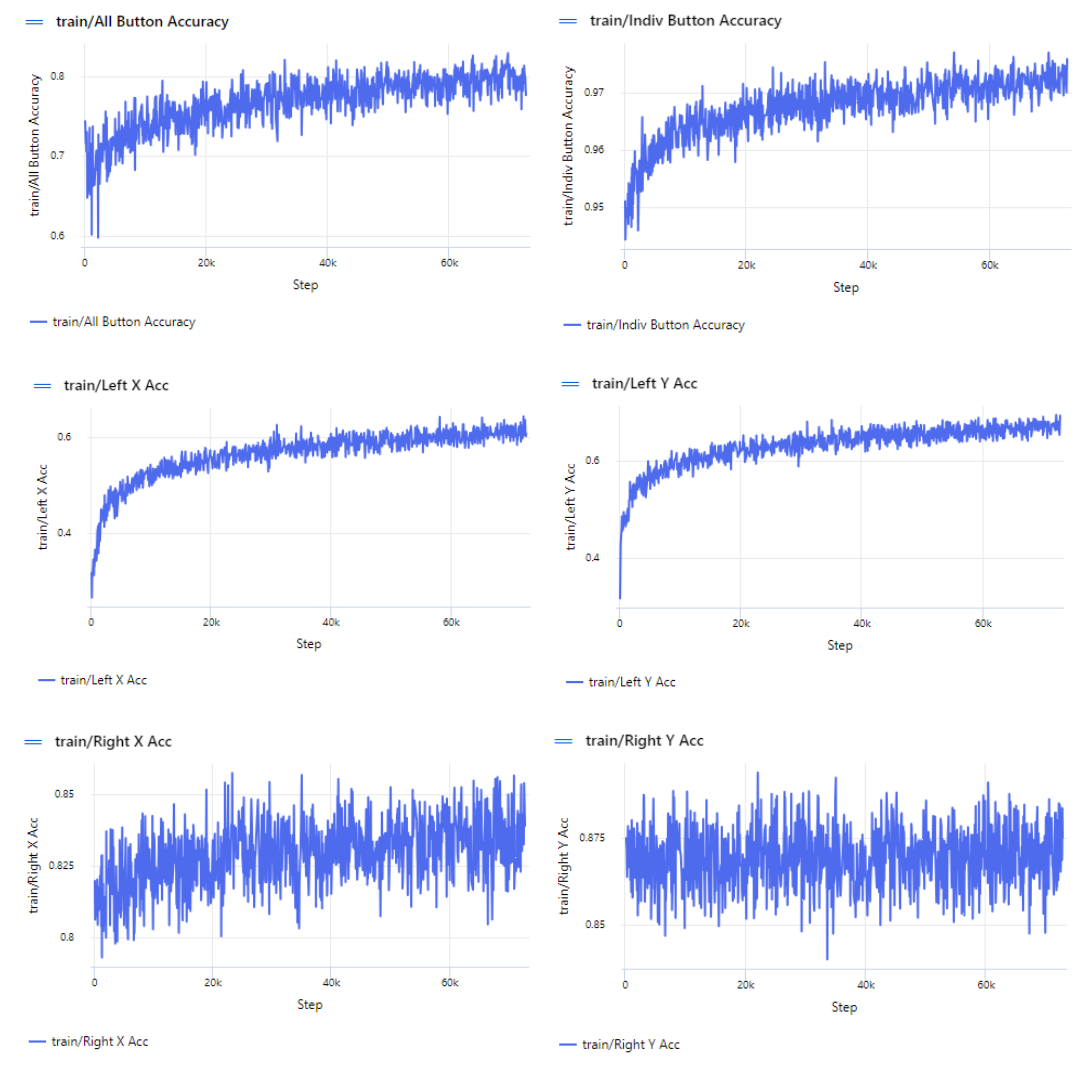}
\caption{\textbf{AI agent action accuracy curves over training}. The Xbox controller action space consists of 12 discrete buttons and two joysticks. \textit{Top:} Button accuracy curves for all the buttons (percentage of times the predicted value of all buttons matches the expected value) on the left; and for individual buttons (mean percentage of times the predicted value of each individual button matches its expected value). \textit{Middle:} Left joystick accuracy curves for $x$ and $y$ components of the joystick. This joystick controls character movement. \textit{Bottom:} Right joystick accuracy curves for $x$ and $y$ components of the joystick. This joystick controls the camera. }
\label{fig-task-sets-list}
\end{center}
\end{figure*}

\begin{figure*}[ht]
\begin{center}
\includegraphics[width=\textwidth]{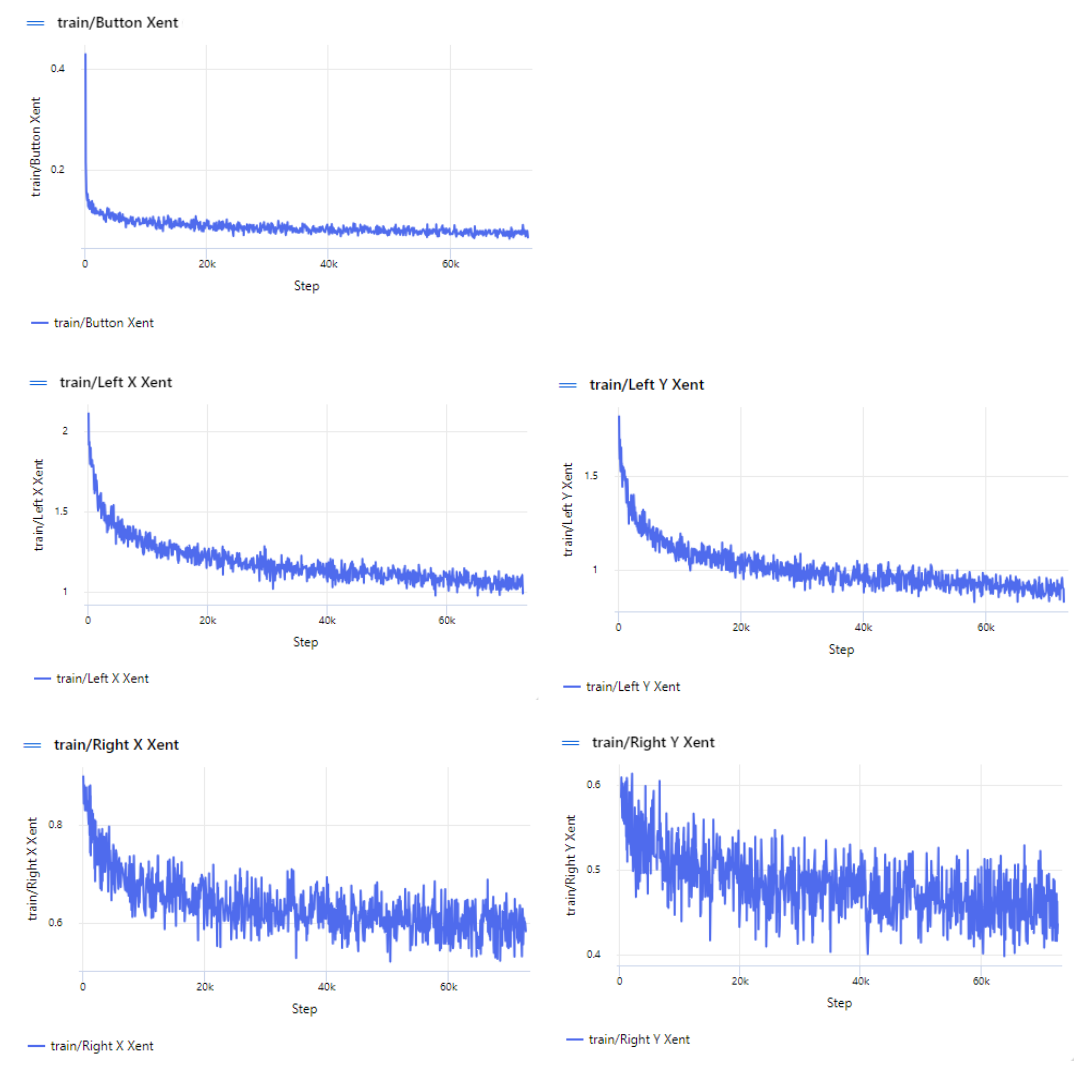}
\caption{\textbf{AI agent decomposed loss curves over training}. The Xbox controller action space consists of 12 discrete buttons and two joysticks with $x$ and $y$ components each of which is discretized into 11 bins. \textit{Top:} Button loss curves for all 12 buttons. \textit{Middle:} Left joystick loss curves for X and Y components of the joystick. This joystick controls character movement. \textit{Bottom:} Right joystick loss curves for X and Y components of the joystick. This joystick controls the camera. }
\label{fig-task-sets-list}
\end{center}
\end{figure*}

%%%%%%%%%%%%%%%%%%%%%%%%%%%%%%%%%%%%%%%%%%%%%%%%%%%%%%%%%%%%%%%%%%%%%%%%%%%%%%%
%%%%%%%%%%%%%%%%%%%%%%%%%%%%%%%%%%%%%%%%%%%%%%%%%%%%%%%%%%%%%%%%%%%%%%%%%%%%%%%

\subsection{Future work, impact and implications}
\label{app:future_impact}
%In future work, we hope to explore the practical utility of the identified behavioral dimensions to investigate how information along these axes can be leveraged for building agents that exhibit specific behaviors. 
\textbf{Future work:} An important direction is to assess how the identified behavioral dimensions can be used practically to develop agents that demonstrate targeted behaviors. 
Our analyses identify different play styles, such as aggressive (fighty) behavior. 
Player data could be separated by play style and used to fine-tune different agents, which may then replicate this behavior more readily.  %For e.g. human behavioral analyses can be used for player-style identification such as identifying aggressive (fighty) players by extracting this data from the behavioral manifold obtained (in Fig.\ref{fig-fight_flight}c). This data can then be used to fine-tune the baseline AI agent on the aggressive player dataset for targeted behavior replication %such as building an aggressive agent. 
A second promising direction is extending the task-sets framework by incorporating automatic discovery and learning of task-sets. 
This would increase the framework's applicability to additionaldomains and make it easier to apply without a preceding substantial data analysis effort. %, but also, make it adaptable and generalizable to diverse domains. 
Finally, a third direction could explore whether any model parameters or components in the latent representations are associable with specific axes of the behavioral manifold. 
This may offer mechanistic insights into the factors influencing high-level behavior, and therefore human-agent alignment.

\textbf{Broader impact:} Our work shows that the alignment of transformer-based models trained on next token prediction may not always be inherent, and it may require specialized training techniques such as supervised fine-tuning \cite{gunel2020supervised, wortsman2022robust, lee2022surgical, kirichenko2022last} and Reinforcement Learning from Human Feedback (RLHF)  \cite{christiano2017deep, stiennon2020learning, glaese2022improving, ouyang2022training, bai2022training, rafailov2023direct, azar2023general}  to achieve alignment. This underscores the importance of our framework for measuring alignment in AI agents. 

\textbf{Broader societal implications:}  Aligning AI with humans along fight-flight responses can help address ethical and moral questions about the use of AI in simulated (and potentially real) conflict situations, defense systems, and decision-making processes that involve risk, uncertainty, and potential harm to individuals and communities. AI alignment with human preferences for solo or multiplayer gameplay can influence social interactions, cultural norms, and community dynamics in gaming and entertainment, shaping how individuals and groups engage with AI-driven gaming experiences, collaborate with AI agents, and form online (and potentially offline) communities. Human-AI alignment along exploration or exploitation can foster innovation, creativity, and adaptive learning in various domains, including research, development, entrepreneurship, and education.

Our framework can contribute to the long-term sustainability of AI technologies by promoting human-centered approaches to AI development and deployment. Since it is an interpretable framework, it can enhance user confidence in AI and promote their widespread adoption.

\subsection{Ethics review program for human data collection}

At Microsoft Research, we have an internal \href{https://www.microsoft.com/en-us/research/microsoft-research-ethics-review-program-irb/}{ethics review program} that helped us work with human data in a way that ensured respect and protection of the rights of human participants contributing to our research.

Our institutional Review Board (IRB) approved all the data collection for the data used in this paper. For the player recordings, we received ethics approval (IRB 10601) from our organization’s Ethics Review Program. Our organization’s Institutional Review Board (IRB) has been officially registered with the U.S. Health and Human Services’ Office for Human Research Protections (OHRP) since 2017.

During data collection, human players playing Bleeding Edge are subjected to a pop up when they log in to the game for the first time to agree to the terms (\href{https://www.microsoft.com/en-gb/servicesagreement}{end-user license agreement}). The behavioral data collected does not contain any personally identifiable information. 

\end{document}